\documentclass[]{main}
\title{\vspace{-1pt}X2SAM: Any Segmentation in Images and Videos}
\author[1,2]{Hao Wang}
\author[3]{Limeng Qiao}
\author[3]{Chi Zhang}
\author[3]{Lin Ma}
\author[3]{Guanglu Wan}
\author[2,\dagger]{Xiangyuan Lan}
\author[1,2,\dagger]{Xiaodan Liang}
\affiliation[1]{Sun Yat-Sen University}
\affiliation[2]{Peng Cheng Laboratory}
\affiliation[3]{Meituan Inc}
\contribution[\dagger]{Corresponding author}
\usepackage{graphicx}
\usepackage{booktabs}
\usepackage{tabularx}
\usepackage{multirow}
\usepackage{diagbox}
\usepackage{hhline}
\usepackage{makecell}
\usepackage{wrapfig}
\usepackage{colortbl}
\usepackage{longtable}
\usepackage{array}
\usepackage{bm}
\usepackage{nicefrac}
\usepackage{changepage}
\usepackage{xspace}
\usepackage{multicol}
\usepackage{url}
\usepackage{titletoc}
\usepackage{siunitx}
\usepackage{nicematrix}
\usepackage{cleveref}
\usepackage{pifont}
\usepackage{amsmath,amsfonts,amssymb}
\usepackage{enumitem}
\setlist[itemize]{leftmargin=*, itemsep=2pt, topsep=3pt}
\setlist[enumerate]{leftmargin=*, itemsep=2pt, topsep=3pt}
\usepackage{algorithm,algpseudocode}

\usepackage[symbol]{footmisc}
\renewcommand{\thefootnote}{\fnsymbol{footnote}}
\definecolor{LightYellow}{HTML}{FEFBE8}   \definecolor{LightGreen}{HTML}{EAF6EE}    \definecolor{LightRed}{HTML}{FDECEA}      \definecolor{LightCyan}{HTML}{E6F6FA}     \definecolor{LightBlue}{HTML}{EAF1FB}     \definecolor{LightIndigo}{HTML}{EEF0FB}   \definecolor{LightGray}{HTML}{F2F4F7}     
\newcommand{\cmark}{\ding{51}}
\newcommand{\xmark}{\ding{55}}
\newcommand{\mj}{$\mathcal{J}$}
\newcommand{\mf}{$\mathcal{F}$}
\newcommand{\mjf}{$\mathcal{J}\&\mathcal{F}$\xspace}

\renewcommand{\paragraph}[1]{\vspace{4pt}\noindent\headbf{#1.}\hspace{0.3em}}
\newcommand{\headbf}[1]{\textcolor{sysugreen}{\textbf{#1}}
}

\newcolumntype{x}[1]{>{\centering\arraybackslash}p{#1pt}}
\newcolumntype{y}[1]{>{\raggedright\arraybackslash}p{#1pt}}
\newcolumntype{z}[1]{>{\raggedleft\arraybackslash}p{#1pt}}
\newcolumntype{P}[1]{>{\centering\arraybackslash}p{#1}}
\newcolumntype{Y}{>{\centering\arraybackslash}X}

\abstract{
    Multimodal Large Language Models (MLLMs) have demonstrated strong image-level visual understanding and reasoning, yet their pixel-level perception across both images and videos remains limited. Foundation segmentation models such as the SAM series produce high-quality masks, but they rely on low-level visual prompts and cannot natively interpret complex conversational instructions. Existing segmentation MLLMs narrow this gap, but are usually specialized for either images or videos and rarely support both textual and visual prompts in one interface. We introduce X2SAM, a unified segmentation MLLM that extends any-segmentation capabilities from images to videos. Given conversational instructions and visual prompts, X2SAM couples an LLM with a Mask Memory module that stores guided vision features for temporally consistent video mask generation. The same formulation supports generic, open-vocabulary, referring, reasoning, grounded conversation generation, interactive, and visual grounded segmentation across image and video inputs. We further introduce the Video Visual Grounded (V-VGD) segmentation benchmark, which evaluates whether a model can segment object tracks in videos from interactive visual prompts. With a unified joint training strategy over heterogeneous image and video datasets, X2SAM delivers strong video segmentation performance, remains competitive on image segmentation benchmarks, and preserves general image and video chat ability.
}
 
\date{\today}
\code{https://github.com/wanghao9610/X2SAM}
\project{https://wanghao9610.github.io/X2SAM}
\correspondence{\email{wanghao9610@gmail.com},\email{xdliang328@gmail.com},\email{lanxy@pcl.ac.cn}}
\begin{document}
\maketitle
\renewcommand{\thefootnote}{\arabic{footnote}}
\setcounter{footnote}{0}
\section{Introduction}

\begin{figure}[ht]
    \centering
    \includegraphics[width=0.98\linewidth]{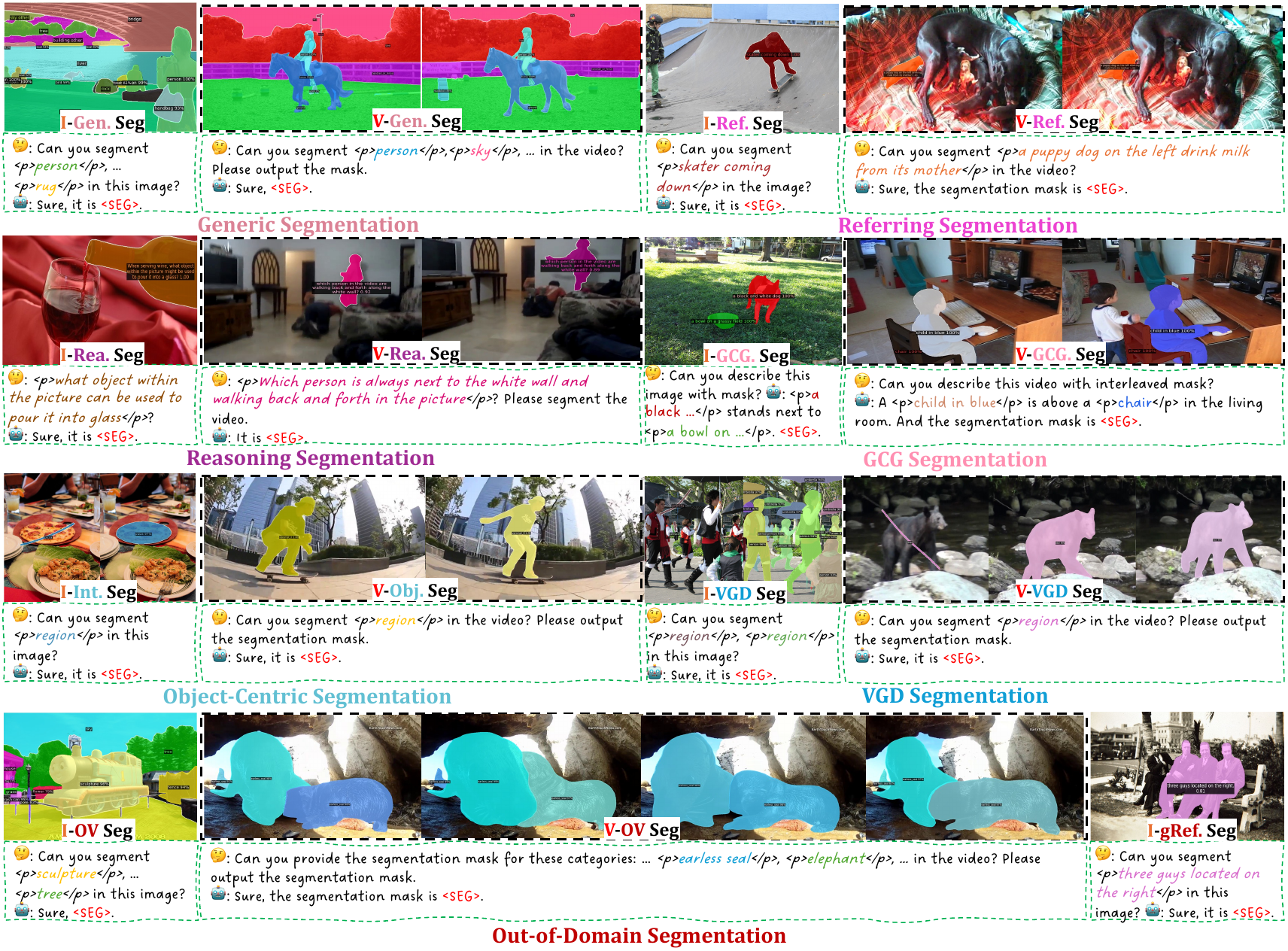}
    \caption{Comprehensive capabilities of X2SAM. Guided by conversational instructions, X2SAM performs diverse \textcolor{orange}{Image} and \textcolor{red}{Video} segmentation tasks. X2SAM provides a unified interface for a wide range of image and video segmentation tasks, including generic, referring, reasoning, grounded conversation generation, interactive, visual grounded, object-centric, and out-of-domain segmentation. This design highlights how image-level any-segmentation capabilities can be extended to video inputs with textual and visual prompts.}
    \label{fig:teaser}
\end{figure}

Multi-modal Large Language Models (MLLMs) have exhibited substantial advancements alongside the rapid development of Large Language Models (LLMs)~\cite{bai2023qwen, touvron2023llama} and multi-modal pre-training methods~\cite{radford2021clip, jia2021align}. These models have shown remarkable effectiveness in a wide range of applications, including image captioning~\cite{xu2015show}, VQA~\cite{antol2015vqa}, and visual editing~\cite{chen2018imgedit}. However, while current MLLMs excel at global visual understanding, their capability to generate dense, pixel-level outputs for precise spatial and temporal comprehension remains limited. This limitation poses a considerable challenge in directly addressing fine-grained tasks across both static images and dynamic video sequences.

\begin{figure}[!tp]
    \centering
    \includegraphics[width=0.98\linewidth]{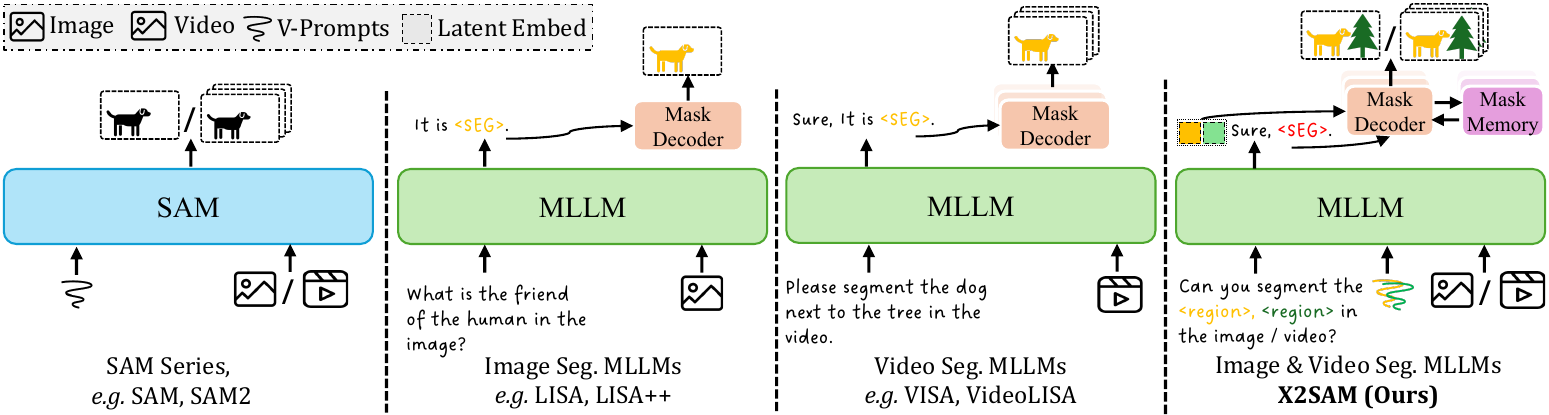}
    \caption{Comparison of X2SAM with existing methods, including SAM Series models, Image Segmentation MLLMs, and Video Segmentation MLLMs. X2SAM processes both textual and visual prompts in a shared image-video segmentation framework, improving coverage over prior image-only or video-only segmentation MLLMs.}
    \label{fig:comparison}
\end{figure}

Foundation segmentation models, such as SAM~\cite{kirillov2023sam} and its video-extended successor SAM2~\cite{ravi2024sam2}, generate dense masks across spatial and temporal domains. Nevertheless, they depend on explicit low-level visual prompts (e.g., points or boxes) and cannot natively interpret complex conversational text instructions. Conversely, as illustrated in \figurename~\ref{fig:comparison}, recent segmentation MLLMs have attempted to bridge language understanding and mask generation, but they remain structurally fragmented. Image segmentation MLLMs (e.g., LISA~\cite{lai2024lisa}) process textual instructions but are restricted to static images and usually lack visual prompting support. Video segmentation MLLMs (e.g., VISA~\cite{yan2024visa}, VideoLISA~\cite{bai2024videolisa}) support temporal text-to-mask generation but do not provide a unified architecture for both static images and visual prompts. Achieving a single framework that interprets complex multi-modal instructions, including both text and visual prompts, for segmentation across images and videos remains a critical challenge.

In this work, we introduce X2SAM, a framework that unifies diverse image and video segmentation tasks and extends the image-centric any-segmentation paradigm toward a unified image-and-video setting. As illustrated in \figurename~\ref{fig:teaser}, X2SAM provides a conversational interface for text-driven and visually prompted segmentation across static images and dynamic videos. To realize this capability and overcome limitations of prior paradigms (\figurename~\ref{fig:comparison}), our approach addresses three technical challenges: (1) \textit{Comprehensive Prompt Integration}: augmenting LLMs to process interleaved textual instructions and visual prompts (V-Prompts) for both image and video inputs. (2) \textit{Spatio-Temporal Task Formulation}: casting diverse image segmentation paradigms into a shared formulation that can represent video targets over time. (3) \textit{Temporal Coherence via Mask Memory}: replacing independent frame-by-frame decoding with a Mask Memory module that interacts with the Mask Decoder and stores guided vision features to maintain mask consistency across video sequences.

As illustrated in \figurename~\ref{fig:framework}, we develop a unified MLLM architecture that processes global visual representations and fine-grained visual features. Guided by latent condition embeddings from the LLM, the Mask Decoder works with the newly introduced Mask Memory module to generate temporally consistent segmentation masks. Moreover, we expand the visual prompting capabilities of MLLMs by introducing the Video Visual Grounded (V-VGD) segmentation task. This task equips the model to segment any instance object in a video using interactive visual prompts, grounding targets across frames.

As shown in \tablename~\ref{tab:capability}, we compare X2SAM with existing methods across inputs, outputs, and tasks. X2SAM is the first to natively support seven segmentation tasks, e.g., generic, open-vocabulary, referring, reasoning, grounded conversation generation, object-centric, and visual grounded segmentation, for image and videos.
 
Supported by a unified joint training strategy that accelerates learning across multi-modalities, X2SAM undergoes co-training with a diverse range of image and video datasets. Experimental results show that X2SAM achieves strong performance across image and video benchmarks, with particularly consistent gains on video segmentation tasks, establishing a practical baseline for unified pixel-level spatio-temporal understanding. In summary, our contributions are as follows:
\begin{itemize}
    \setlength{\itemsep}{2pt}
    \item We introduce X2SAM, a unified framework that extends the any segmentation paradigm from images to videos. By integrating an MLLM with a Mask Memory module, X2SAM formulates diverse image and video segmentation tasks into a standardized, temporally consistent format.
    \item We propose a new benchmark, Video Visual Grounded (V-VGD) segmentation, which provides interactive visual prompts for MLLMs to ground and segment instance objects consistently across video frames.
    \item We present a unified joint training strategy to co-train X2SAM on both image and video data. Extensive evaluations show that X2SAM supports a broad set of segmentation tasks, remains competitive on image benchmarks, and achieves strong results on video and out-of-domain evaluations.
\end{itemize}

\section{Related Work}
\noindent\headbf{Multi-modal Large Language Model.} Multi-modal learning has witnessed progressive developments alongside the rapid evolution of Large Language Models (LLMs)~\cite{bai2023qwen,touvron2023llama} and multi-modal pre-training methods~\cite{radford2021clip, jia2021align}. The field has evolved from early models focused on task-specific fusion and feature extraction~\cite{li2022blip}, to generalized, instruction-tuned frameworks leveraging visual feature tokenization~\cite{liu2023visual, liu2024llava1x5}. While current MLLMs demonstrate remarkable effectiveness in global visual understanding tasks such as image captioning~\cite{xu2015show} and VQA~\cite{antol2015vqa}, their capability to generate dense, pixel-level outputs for precise spatial and temporal comprehension remains highly limited. This poses a considerable challenge when directly addressing fine-grained tasks across static images and dynamic video sequences.

\noindent\headbf{Image Segmentation MLLMs.} Foundation models like SAM~\cite{kirillov2023sam} and its extensions~\cite{ravi2024sam2} have profoundly impacted the segmentation landscape by introducing visual grounding signals, vastly improving mask generation performance. Building upon this, researchers have explored combining MLLMs with segmentation models to handle open-world challenges, unified task architectures~\cite{athar2023tarvis, jain2023oneformer}, and language-guided tasks~\cite{li2024omgseg,zhang2024omgllava}. Image segmentation MLLMs, such as LISA~\cite{lai2024lisa}, successfully process complex textual instructions to output segmentation masks. However, these models are structurally restricted to static images and frequently lack comprehensive support for interactive visual prompting (V-Prompts), limiting their ability to treat grounded visual inputs as freely as textual inputs.

\noindent\headbf{Video Segmentation MLLMs.} Extending dense segmentation capabilities to dynamic video sequences introduces significant temporal complexities~\cite{wang2021tmanet,li2022videoknet}. Recent video segmentation MLLMs, including VISA~\cite{yan2024visa} and VideoLISA~\cite{bai2024videolisa}, have attempted to bridge this gap by enabling temporal text-to-mask generation. Despite these advancements, the current landscape remains structurally fragmented. Existing video-centric MLLMs lack the unified architecture for both images and videos. Furthermore, standard frame-by-frame decoding approaches struggle to systematically store and track multi-modal guided features, failing to maintain robust mask consistency and temporal coherence across continuous video frames.

\noindent\headbf{Analysis against SAM2 and X-SAM.}
X2SAM is related to SAM2~\cite{ravi2024sam2} and X-SAM~\cite{wang2026xsam}, but targets a distinct setting. SAM2 enables promptable image and video segmentation with memory-based propagation, yet it mainly relies on low-level visual prompts and lacks language-driven reasoning or grounded conversation. X-SAM supports MLLM-based segmentation with textual and visual prompts, but is image-centric and does not model temporal object identity.
X2SAM is not a simple X-SAM+SAM2 cascade. It unifies image and video segmentation in an instruction-following framework, where textual prompts, visual prompts, and generated \texttt{<SEG>} tokens are converted into mask-aware conditions. Its language-conditioned Mask Memory stores guided visual features from the MLLM-conditioned decoder, coupling semantic grounding with temporal propagation. Thus, unlike frame-wise X-SAM or cascaded propagation, X2SAM jointly optimizes grounding, decoding, and memory for temporally consistent instruction-based mask generation.
\begin{table*}[!tp]
    \centering
    \setlength{\tabcolsep}{2pt}
    \caption{Comparison of Chat-based and Segmentation-based MLLMs across inputs, outputs, and tasks.}
    \label{tab:capability}
    \resizebox{\textwidth}{!}{
        \begin{tabular}{l cccc cc cccc}
            \toprule
            \multirow{2}{*}{\textbf{Method}}                       & \multicolumn{4}{c}{\textbf{Inputs}} & \multicolumn{2}{c}{\textbf{Outputs}} & \multicolumn{4}{c}{\textbf{Tasks}}                                                                                                                                                                                                          \\
            \cmidrule(lr){2-5} \cmidrule(lr){6-7} \cmidrule(lr){8-11}
                                                                   & \textbf{\textit{Image}}             & \textbf{\textit{Video}}              & \textbf{\textit{T-Prompts}}        & \textbf{\textit{V-Prompts}}
                                                                   & \textbf{\textit{Text}}              & \textbf{\textit{Mask}}
                                                                   & \textbf{\textit{Image-Chat}}        & \textbf{\textit{Video-Chat}}         & \textbf{\textit{\#Image-Seg.}}     & \textbf{\textit{\#Video-Seg.}}                                                                                                                                                                         \\
            \midrule
\rowcolor{gray!15} \multicolumn{11}{l}{\textbf{\textit{Chat-based MLLMs}}}                                                                                                                                                                                                                                                                                                        \\
            \textcolor{gray}{LLaVA~\cite{liu2023llava}}            & \textcolor{gray}{\cmark}            & \textcolor{gray}{\xmark}             & \textcolor{gray}{\xmark}           & \textcolor{gray}{\xmark}       & \textcolor{gray}{\cmark} & \textcolor{gray}{\xmark} & \textcolor{gray}{\cmark} & \textcolor{gray}{\xmark} & \textcolor{gray}{0}         & \textcolor{gray}{0}         \\
            \textcolor{gray}{LLaVA-Next~\cite{liu2024llavanext}}   & \textcolor{gray}{\cmark}            & \textcolor{gray}{\cmark}             & \textcolor{gray}{\cmark}           & \textcolor{gray}{\xmark}       & \textcolor{gray}{\cmark} & \textcolor{gray}{\xmark} & \textcolor{gray}{\cmark} & \textcolor{gray}{\cmark} & \textcolor{gray}{0}         & \textcolor{gray}{0}         \\
            \textcolor{gray}{LLaVA-OV~\cite{li2024llavaov}}        & \textcolor{gray}{\cmark}            & \textcolor{gray}{\cmark}             & \textcolor{gray}{\cmark}           & \textcolor{gray}{\xmark}       & \textcolor{gray}{\cmark} & \textcolor{gray}{\xmark} & \textcolor{gray}{\cmark} & \textcolor{gray}{\cmark} & \textcolor{gray}{0}         & \textcolor{gray}{0}         \\
            \textcolor{gray}{Intern-VL~\cite{chen2024internvl1.5}} & \textcolor{gray}{\cmark}            & \textcolor{gray}{\cmark}             & \textcolor{gray}{\cmark}           & \textcolor{gray}{\xmark}       & \textcolor{gray}{\cmark} & \textcolor{gray}{\xmark} & \textcolor{gray}{\cmark} & \textcolor{gray}{\cmark} & \textcolor{gray}{0}         & \textcolor{gray}{0}         \\
            \textcolor{gray}{Qwen-VL~\cite{qwenvl}}                & \textcolor{gray}{\cmark}            & \textcolor{gray}{\cmark}             & \textcolor{gray}{\cmark}           & \textcolor{gray}{\xmark}       & \textcolor{gray}{\cmark} & \textcolor{gray}{\xmark} & \textcolor{gray}{\cmark} & \textcolor{gray}{\cmark} & \textcolor{gray}{0}         & \textcolor{gray}{0}         \\
            \midrule
\rowcolor{gray!15} \multicolumn{11}{l}{\textbf{\textit{Seg.-based MLLMs}}}                                                                                            
                                                                                                            \\
            LISA~\cite{lai2024lisa}                                & \cmark                              & \textcolor{gray}{\xmark}                               & \cmark                             & \textcolor{gray}{\xmark}       & \cmark                   & \cmark                   & \cmark                   & \textcolor{gray}{\xmark} & 2                           & 2                           \\
            VISA~\cite{yan2024visa}                                & \cmark                              & \cmark                               & \cmark                             & \textcolor{gray}{\xmark}       & \cmark                   & \cmark                   & \cmark                   & \cmark                   & 2                           & 2                           \\
            GLaMM~\cite{rasheed2024glamm}                          & \cmark                              & \textcolor{gray}{\xmark}             & \cmark                             & \textcolor{gray}{\xmark}       & \cmark                   & \cmark                   & \textcolor{gray}{\xmark} & \cmark                   & 2                           & 0                           \\
            VideoGLaMM~\cite{munasinghe2024videoglamm}             & \textcolor{gray}{\xmark}            & \cmark                               & \cmark                             & \textcolor{gray}{\xmark}       & \cmark                   & \cmark                   & \textcolor{gray}{\xmark} & \cmark                   & 0                           & 2                           \\
            PSALM~\cite{zhang2024psalm}                            & \cmark                              & \cmark                               & \cmark                             & \cmark                         & \cmark                   & \cmark                   & \cmark                   & \textcolor{gray}{\xmark} & 5                           & 1                           \\
            HyperSeg~\cite{wei2024hyperseg}                        & \cmark                              & \cmark                               & \cmark                             & \cmark                         & \cmark                   & \cmark                   & \cmark                   & \textcolor{gray}{\xmark} & 5                           & 4                           \\
            Sa2VA~\cite{yuan2025sa2va}                             & \cmark                              & \cmark                               & \cmark                             & \cmark                         & \cmark                   & \cmark                   & \cmark                   & \cmark                   & 3                           & 3                           \\
            X-SAM~\cite{wang2026xsam}                              & \cmark                              & \cmark                               & \cmark                             & \cmark                         & \cmark                   & \cmark                   & \cmark                   & \textcolor{gray}{\xmark}                   & \textbf{7} & 0                           \\
\rowcolor{LightGreen} \textbf{X2SAM}                    & \textbf{\cmark}                     & \textbf{\cmark}                      & \textbf{\cmark}                    & \textbf{\cmark}                & \textbf{\cmark}          & \textbf{\cmark}          & \textbf{\cmark}          & \textbf{\cmark}          & \textbf{7} & \textbf{7} \\
            \bottomrule
        \end{tabular}
    }
\end{table*} 

\section{Method}
To extend segmentation capabilities seamlessly from static images to dynamic video sequences, we propose a novel segmentation-oriented MLLM, termed X2SAM. We first present the formal problem formulation of X2SAM, encompassing the definition of inputs, outputs, and task formulations. Subsequently, we elaborate on the architectural framework of X2SAM, detailing the input processing pipeline, the encoders and LLM, the redesigned mask decoder, and the mask memory module. Finally, we discuss the training methodology of X2SAM, highlighting our unified joint training strategy and the associated training objectives.

\subsection{Formulation}
\noindent\headbf{Inputs.} The inputs to X2SAM comprise a textual or visual prompt coupled with either a single image or a video sequence. The textual prompt constitutes a natural language instruction that delineates the target segmentation task, whereas the visual prompt represents an interactive visual cue (e.g., points or boxes) that designates the objects of interest. The image or video sequence serves as the primary visual input to be processed by the framework.

\noindent\headbf{Outputs.} The outputs of X2SAM comprise a contextual language response and a corresponding segmentation mask. The language response represents the natural language output generated by the LLM, while the segmentation mask provides a binary, pixel-level delineation of the target specified by the prompt.

\noindent\headbf{Unified Formulation.} To accommodate a comprehensive set of image and video segmentation tasks, we introduce a unified formulation for X2SAM. In this formulation, the objects of interest across all tasks are treated as conditional states, while the language instruction serves as the contextual input. Following X-SAM~\cite{wang2026xsam}, we incorporate two special tokens, \texttt{<p>} and \texttt{</p>}, to demarcate the beginning and end of the object condition, respectively, along with a dedicated \texttt{<SEG>} token to indicate the corresponding segmentation mask. The LLM's output representation for the \texttt{<SEG>} token functions as a dedicated directive, guiding the mask decoder to segment the objects of interest. Furthermore, task-specific templates are devised to facilitate aligned language response generation by the LLM.

\subsection{Framework}
\begin{figure}[!tp]
    \centering
    \includegraphics[width=0.98\linewidth]{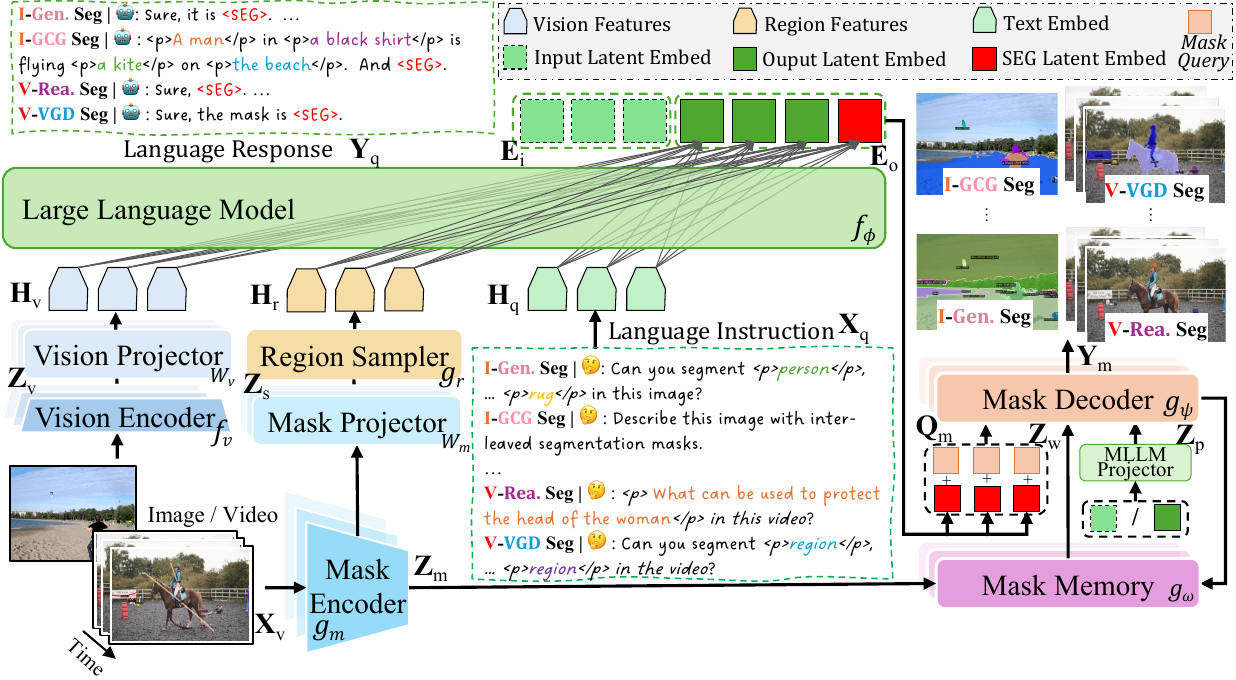}
    \caption{Overview of X2SAM. The \textrm{Vision Encoder} extracts global visual representations, while the \textrm{Mask Encoder} captures fine-grained visual features. The \textrm{Large Language Model} generates the language response and produces the latent condition embedding, which guides the \textrm{Mask Decoder} in generating the segmentation mask. The \textrm{Mask Memory} module stores guided vision features for each video frame, and the \textrm{Region Sampler} extracts region-of-interest embeddings from both images and videos.}
    \label{fig:framework}
\end{figure}

 \noindent\headbf{Overview.} As illustrated in \figurename~\ref{fig:framework}, X2SAM takes as input a language instruction $\textrm{X}_\textrm{q}$ and a visual input $\textrm{X}_\textrm{v} \in \mathbb{R}^{T \times H \times W \times C}$, where $T=1$ for images and $T>1$ for videos, and jointly outputs a language response $\textrm{Y}_\textrm{q}$ and a segmentation mask $\textrm{Y}_\textrm{m} \in \mathbb{R}^{T \times H \times W}$. The model adopts a dual-branch visual extraction architecture: a vision encoder $f_v$ extracts global representations $\textrm{Z}_\textrm{v}$, while a mask encoder $g_m$ captures fine-grained features $\textrm{Z}_\textrm{m}$ for dense prediction. The projected global features $\textrm{H}_\textrm{v} = \mathbf{\textit{W}}_\textrm{v}(\textrm{Z}_\textrm{v})$, region features $\textrm{H}_\textrm{r}$ from the region sampler $g_r$, and tokenized textual embeddings $\textrm{H}_\textrm{q}$ are fed into the LLM $f_\phi$.
The LLM auto-regressively generates $\textrm{Y}_\textrm{q}$ together with a dedicated \texttt{SEG} latent embedding, serving as a semantic bridge between language understanding and mask prediction. This embedding is transformed by the MLLM projector into the prompt token embedding $\textrm{Z}_\textrm{p}$. Finally, the mask decoder $g_\psi$ synthesizes $\textrm{Y}_\textrm{m}$ by integrating $\textrm{Z}_\textrm{p}$, learnable mask queries $\textrm{Q}_\textrm{m}$, and temporally refined visual features $\textrm{Z}_\textrm{w}$. These features are produced by the mask memory module $g_\omega$, which maintains a first-in-first-out (FIFO) cache of guided visual features from preceding frames for temporally consistent segmentation.

\noindent\headbf{Input Processing.} Given the visual input $\textrm{X}_\textrm{v}$ and instruction $\textrm{X}_\textrm{q}$, X2SAM employs two complementary visual processing pipelines. For global understanding, we follow Qwen3-VL-4B~\cite{qwen3vl}, where visual inputs are augmented with timestamps, partitioned into spatial patches, and projected into latent embeddings $\textrm{Z}_\textrm{v}$. For high-resolution mask prediction, we adopt SAM2~\cite{ravi2024sam2}, which processes videos frame-wise to extract fine-grained mask features $\textrm{Z}_\textrm{m}$. When region-specific information is required, the region sampler $g_r$ extracts localized visual prompt embeddings from $\textrm{Z}_\textrm{m}$. In parallel, the textual instruction is formatted with task-specific templates, tokenized, and embedded into text latent representations $\textrm{H}_\textrm{q}$.

\noindent\headbf{Vision Encoder and LLM.} Large Vision-Language Models (LVLMs) inherently possess robust semantic understanding. We adopt the vision encoder, vision projector, and LLM backbone from Qwen3-VL~\cite{qwen3vl}, endowing X2SAM with state-of-the-art multimodal reasoning and broad visual comprehension capabilities.

\noindent\headbf{Region Sampler.} We design a parameter-free region sampler to facilitate the injection of visual prompts into the LLM. Specifically, we conduct point-sampling~\cite{you2023ferret} on regions of interest utilizing the mask encoder's high-resolution features $\textrm{Z}_\textrm{m}$. We then apply adaptive pooling to aggregate these point-sampled features into cohesive region-level representations $\textrm{H}_\textrm{r}$.

\begin{figure}[!tp]
    \centering
    \includegraphics[width=0.98\linewidth]{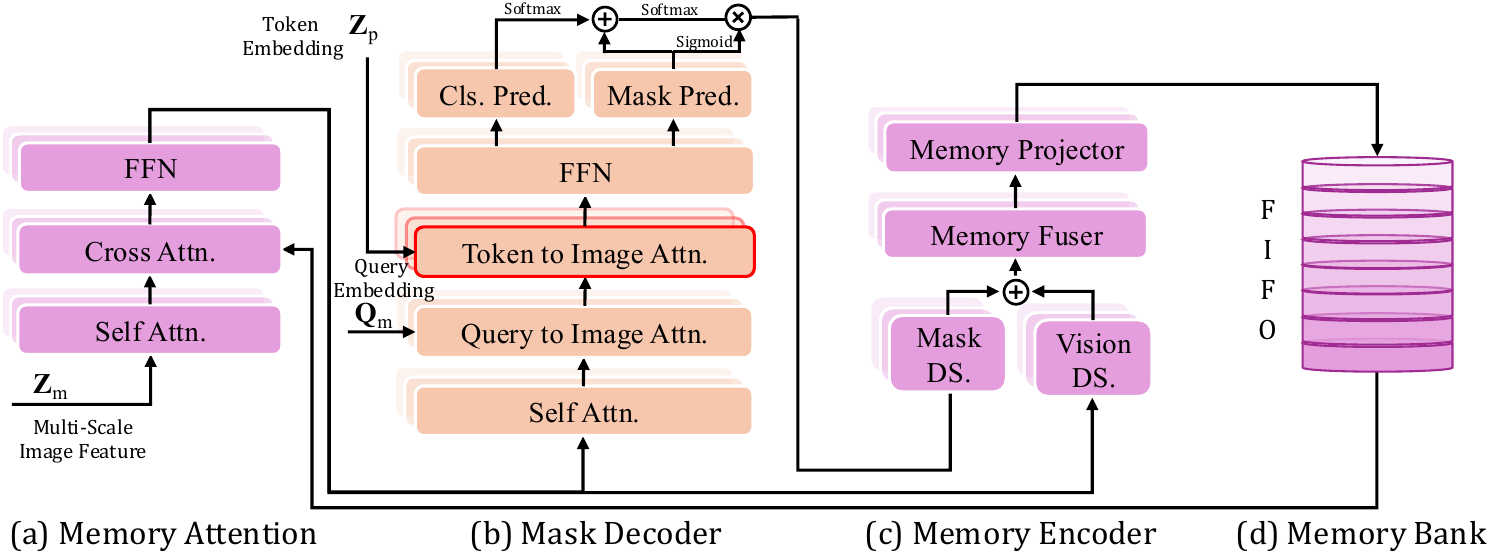}
    \caption{Architecture of the mask memory and mask decoder. (a) \textit{Memory Attention} attends the guided vision features of previous frames in the video. (b) \textit{Mask Decoder} generates the segmentation mask of the current frame. (c) \textit{Memory Encoder} encodes the downsampled vision features and mask logits of the current frame. (d) \textit{Memory Bank} stores the guided vision features of each frame in the video, and updates the memory bank via FIFO (First In First Out) strategy.}
    \label{fig:memory}
\end{figure}

\noindent\headbf{Mask Encoder and Decoder.} We utilize the robust and lightweight mask encoder from SAM2~\cite{ravi2024sam2}. However, to overcome limitations in parallel mask generation, we discard its original mask decoder and redesign a novel architecture inspired by X-SAM~\cite{wang2026xsam}. As illustrated in \figurename~\ref{fig:memory}(b), we introduce structured attention modules, namely \textit{Query-to-Image Attention} and \textit{Token-to-Image Attention}, to inject token-level conditional information into the mask decoder. This allows the LLM's semantic token embedding $\textrm{Z}_\textrm{p}$ to directly interact with spatial features. We employ zero-initialization for the Token-to-Image Attention parameters, ensuring smooth and stable integration of token-level conditional information during early training.

 \noindent\headbf{Mask Memory.} To maintain temporal coherence across video frames, we propose a Mask Memory module (detailed in \figurename~\ref{fig:memory}) that operates as a temporal cache. Its data flow follows the four parts in \figurename~\ref{fig:memory}:
1) \textit{Memory Attention} (\figurename~\ref{fig:memory}a): attends to guided vision features from previous frames and produces temporally-refined vision features for the current frame.
2) \textit{Mask Decoder} (\figurename~\ref{fig:memory}b): generates the current-frame segmentation mask and mask logits from temporally-refined features and the LLM-derived segmentation token.
3) \textit{Memory Encoder} (\figurename~\ref{fig:memory}c): encodes downsampled vision features and current-frame mask logits into guided vision features.
4) \textit{Memory Bank} (\figurename~\ref{fig:memory}d): stores guided vision features of processed frames and updates the memory bank using a First-In-First-Out (FIFO) strategy.

\subsection{Training}
\label{sec:training}
\noindent\headbf{Agnostic Segmentor Training.} We first perform category-agnostic segmentor training to provide the mask decoder with a stable initialization before multimodal instruction tuning. Following X-SAM~\cite{wang2026xsam}, the mask encoder is kept frozen and only the mask decoder is optimized with mask-level supervision. This stage encourages the decoder to learn class-independent shape and boundary priors from dense annotations, thereby reducing its dependence on semantic category labels during the subsequent joint training stage.
Our mask loss $\mathcal{L}_{\mathrm{mask}}$ combines binary cross-entropy loss $\mathcal{L}_{\mathrm{bce}}$ and dice loss $\mathcal{L}_{\mathrm{dice}}$~\cite{milletari2016diceloss}:
\begin{equation}
    \mathcal{L}_{\mathrm{mask}} = \lambda_{\mathrm{bce}} \mathcal{L}_{\mathrm{bce}} + \lambda_{\mathrm{dice}} \mathcal{L}_{\mathrm{dice}},
\end{equation}
where $\lambda_{\mathrm{bce}} = 5.0$ and $\lambda_{\mathrm{dice}} = 5.0$ balance the relative weighting of each objective.

\noindent\headbf{Unified Joint Training.} We train X2SAM jointly on heterogeneous image and video datasets under a unified optimization framework. The main challenge is that image and video samples differ substantially in temporal length and memory footprint. To address this issue, we adopt a dimension-shifting pipeline together with modality-aware batching. Given a visual input tensor $\mathbf{X}_\textrm{v} \in \mathbb{R}^{B \times T \times H \times W \times 3}$, where $T=1$ for images and $T>1$ for videos, we first transpose it to $\mathbb{R}^{T \times B \times H \times W \times 3}$ and split it into $T$ frame-level tensors of shape $\mathbb{R}^{B \times H \times W \times 3}$. Each frame is then processed by the mask encoder using the same image-level interface, while temporal dependencies are introduced through the mask memory module during sequential mask decoding. The predicted frame-level masks are finally concatenated along the temporal dimension to recover the sequence-level output $\mathbb{R}^{B \times T \times H \times W}$.
To improve training efficiency under memory constraints, we further adapt the batch organization to the input modality. We set the base per-device batch size to $B=1$ for video samples to avoid excessive memory consumption, while image-only batches are expanded with an image batch multiplier, yielding an effective image batch size on per device to better utilize GPU parallelism. We also use modality-specific gradient accumulation, updating image batches every step and accumulating video gradients over multi-steps to stabilize optimization under the same memory budget. In addition, a temporal-aware sampler groups video clips with the same temporal length into the same batch, reducing unnecessary padding and improving computational efficiency. 
Our joint training objective $\mathcal{L}_{\mathrm{joint}}$ integrates the auto-regressive loss $\mathcal{L}_{\mathrm{ar}}$~\cite{radford2018gpt1} for language generation, the mask loss $\mathcal{L}_{\mathrm{mask}}$ for mask segmentation, and the focal loss $\mathcal{L}_{\mathrm{cls}}$~\cite{lin2017focalloss} for mask classification:
\begin{equation}
    \mathcal{L}_{\mathrm{joint}} =
    \begin{cases}
        \mathcal{L}_{\mathrm{ar}},                                                            & \mathrm{image~\&~video~chat}         \\
        \mathcal{L}_{\mathrm{ar}} + \mathcal{L}_{\mathrm{mask}} + \mathcal{L}_{\mathrm{cls}}, & \mathrm{image~\&~video~segmentation}
    \end{cases}
\end{equation}
 \section{Experiments}
\subsection{Tasks, Datasets, and Metrics}
\label{sec:task_datas_metrics}
\noindent\headbf{Tasks.} X2SAM is engineered to perform segmentation across both static images and video sequences, driven by textual or visual prompts. The framework spans a comprehensive suite of 14 segmentation tasks, stratified into image-based and video-based modalities.
The seven image-based tasks include: generic segmentation (I-Gen.), open-vocabulary segmentation (I-OV), referring segmentation (I-Ref.), reasoning segmentation (I-Rea.), grounded conversation generation segmentation (I-GCG), image interactive segmentation (I-Int.), and visual grounded segmentation (I-VGD).
Correspondingly, the seven video-based tasks comprise: generic segmentation (V-Gen.), open-vocabulary segmentation (V-OV), referring segmentation (V-Ref.), reasoning segmentation (V-Rea.), grounded conversation generation segmentation (V-GCG), video object segmentation (V-Obj.), and visual grounded segmentation (V-VGD).

\noindent\headbf{Datasets.} Our training has two phases: class-agnostic segmentor training and unified joint training. For the agnostic segmentor phase, we use mask-only SA-1B~\cite{kirillov2023sam} to train the mask decoder. The unified joint training phase integrates data from all 14 segmentation tasks, supplemented by image and video chat datasets.
For image segmentation and chat tasks, we follow the mixed fine-tuning configuration of X-SAM~\cite{wang2026xsam}. The video segmentation corpus includes: VIPSeg~\cite{miao2022vipseg}, VSPW~\cite{miao2021vspw}, and YT-VIS19~\cite{yang2019ytvis} for generic segmentation; YT-RefVOS~\cite{seo2020urvos} for referring segmentation; ReVOS~\cite{yan2024visa} for reasoning segmentation; VideoGLaMM~\cite{munasinghe2024videoglamm} for grounded conversation generation; and YT-VOS19~\cite{xu2018ytvos} and DAVIS17~\cite{perazzi2016davis} for video object segmentation.
We introduce two datasets for video visual grounded segmentation, derived from YT-VIS19 and VIPSeg. For video chat training, we use VideoInstruct100K~\cite{maaz2024videochatgpt}.
To assess generalization, we evaluate on standard validation or test splits and out-of-domain benchmarks: gRefCOCO~\cite{liu2023gres} for image referring segmentation, ADE20K~\cite{zhou2019ade20k} for image open-vocabulary segmentation, and YT-VIS-21~\cite{yang2019ytvis} for video open-vocabulary segmentation.

\noindent\headbf{Metrics.} We evaluate X2SAM across image and video benchmarks, following established evaluation protocols.
For panoptic, instance, and semantic segmentation tasks in both images and videos, we report (V)PQ, mAP, and mIoU, respectively. Image referring and reasoning segmentation are evaluated using cIoU and gIoU. Video-based referring, reasoning, and video object segmentation tasks are evaluated using \mj, \mf, and~\mjf.
For GCG segmentation, we report METEOR, CIDEr, AP50, and mIoU, and VGD segmentation performance is quantified using mAP and AP50. Image interactive segmentation utilizes mIoU and cIoU.

\subsection{Implementation Details}
\noindent\headbf{Baseline Setup.} We initialize the vision encoder, vision projector, and LLM with pre-trained weights from Qwen3-VL~\cite{qwen3vl}, while the mask encoder is initialized from SAM2~\cite{ravi2024sam2}. We employ LoRA~\cite{hu2022lora} for LLM fine-tuning. The mask decoder is initialized using the pre-trained agnostic segmentor. Unless otherwise specified, the region sampler uses mask-encoder features with an adaptive pooling kernel size of 4. The ablation baseline omits the token-to-image attention layers in the mask decoder and excludes the mask memory module; the full X2SAM model adds both components. Both the mask and MLLM projectors are implemented as MLPs.
\noindent\headbf{Training Setup.} In the agnostic segmentor training phase, the mask encoder is frozen and the mask decoder is trained with an effective batch size of 128 and a learning rate of $1\times10^{-4}$. In the unified joint training phase, we optimize the projectors, LoRA parameters of the LLM, mask encoder, mask decoder, and mask memory. The learning rate is $1\times10^{-5}$ for the mask encoder and $1\times10^{-4}$ for the other trainable modules. The effective batch size is 32 for video data and 128 for image data. The memory capacity in the mask memory module is fixed at 8 frames unless the memory-size ablation states otherwise. Optimization is performed using AdamW~\cite{loshchilov2017adamw} with a weight decay of 0.05. To ensure balanced representation across heterogeneous sources, we apply a dataset-balanced resampling strategy~\cite{wang2026xsam} with a temperature parameter $t=0.1$. All main experiments are trained for one epoch on 32 NVIDIA H800 GPUs, while ablation studies are conducted on 16 GPUs due to limited computational resources.

\subsection{Ablation Studies}
\begin{table*}[!tp]
    \centering
    \begin{minipage}[t]{0.46\textwidth}
        \centering
        \setlength{\tabcolsep}{2pt}
        \caption{Ablation on the mask decoder.}
        \resizebox{\textwidth}{!}{
            \begin{tabular}{lcccc}
                \toprule
                \multirow{3}{*}{\textbf{Method}}
                 & \textbf{I-Ref.}                              & \textbf{I-Rea.}                   & \textbf{V-Ref.}  & \textbf{V-Rea.}                                    \\
                 & \scriptsize RefCOCO/+/g                      & \scriptsize Val/Test              & \scriptsize YT21 & \scriptsize Ref./Rea./All                          \\
                 & \scriptsize cIoU                             & \scriptsize Val/Test              & \scriptsize \mjf & \scriptsize \mjf                                   \\
                \midrule
                Baseline
                 & \underline{82.9}/\textbf{78.0}/\textbf{79.5} & 58.6/55.9                         & 53.6             & 36.3/36.9/36.5                                     \\
                + T2I(Rand)
                 & 82.5/77.2/\underline{79.4}                   & \underline{62.4}/\underline{56.0} & \underline{58.2} & \underline{44.9}/\underline{45.7}/\underline{44.9} \\
                + T2I(Zero)
                 & \textbf{83.3}/\underline{77.8}/\textbf{79.5} & \textbf{63.0}/\textbf{56.8}       & \textbf{60.8}    & \textbf{47.6}/\textbf{47.7}/\textbf{47.3}          \\
                \bottomrule
            \end{tabular}
            \label{tab:maskdec}
        }
    \end{minipage}\hfill
    \begin{minipage}[t]{0.52\textwidth}
        \centering
        \setlength{\tabcolsep}{2pt}
        \caption{Ablation on the joint training methods.}
        \resizebox{\textwidth}{!}{
            \begin{tabular}{lcccccc}
                \toprule
                \multirow{3}{*}{\textbf{Method}}
                 & \textbf{Train}                                     & \textbf{I-Gen.}            & \textbf{I-OV}              & \textbf{V-Gen.}            & \textbf{V-OV}    \\
                 & \scriptsize Time                                   & \scriptsize Pan./Sem./Ins. & \scriptsize Pan./Sem./Ins. & \scriptsize Pan./Sem./Ins. & \scriptsize Ins. \\
                 & \scriptsize Hours
                 & \scriptsize PQ/mIoU/mAP                            & \scriptsize PQ/mIoU/mAP    & \scriptsize VPQ/mIoU/mAP   & \scriptsize mAP                               \\
                \midrule
                Separate
                 & /
                 & 52.6/64.0/43.8
                 & 25.3/32.5/19.5
                 & 42.9/61.1/66.3
                 & 57.1                                                                                                                                                         \\
                Simple
                 & $\sim$\underline{5.2K}
                 & \textbf{54.4}/\underline{65.0}/\textbf{45.0}
                 & \underline{29.3}/\underline{36.3}/\underline{18.9}
                 & \underline{46.8}/\underline{64.6}/\underline{68.1}
                 & \underline{58.7}                                                                                                                                             \\
                Unified
                 & \textbf{3.3K}
                 & \underline{54.1}/\textbf{65.3}/\underline{44.8}
                 & \textbf{29.4}/\textbf{37.5}/\textbf{19.3}
                 & \textbf{47.1}/\textbf{64.7}/\textbf{68.3}
                 & \textbf{59.1}                                                                                                                                                \\
                \bottomrule
            \end{tabular}
            \label{tab:training}
        }
    \end{minipage}
\end{table*}
 \begin{table*}[tp]
    \centering
    \begin{minipage}[t]{0.48\textwidth}
        \centering
        \setlength{\tabcolsep}{2pt}
        \caption{Ablation on the mask memory.}
        \resizebox{\textwidth}{!}{
            \begin{tabular}{lccc}
                \toprule
                \multirow{3}{*}{\textbf{Method}}
                 & \textbf{V-Gen.}                                    & \textbf{V-Ref.}       & \textbf{V-Rea.}           \\
                 & \scriptsize Pan./Sem./Ins.                         & \scriptsize YT21/DV17 & \scriptsize Ref./Rea./All \\
                 & \scriptsize VPQ/mIoU/mAP                           & \scriptsize \mjf      & \scriptsize \mjf          \\
                \midrule
                Baseline
                 & 42.9/61.1/66.3
                 & 53.6/41.1
                 & 36.3/36.9/36.5                                                                                         \\
                +Single Scale
                 & 42.7/60.8/66.1
                 & 52.5/41.9
                 & 38.0/37.5/37.6                                                                                         \\
                +Mask Guide
                 & 44.5/\underline{62.3}/\underline{66.7}
                 & 63.3/\underline{49.4}
                 & 51.4/51.4/51.1                                                                                         \\
                +Class Guide
                 & \underline{44.8}/61.9/\textbf{66.8}
                 & \underline{64.6}/48.3
                 & \underline{52.0}/\underline{51.9}/\underline{51.6}                                                     \\
                +Multi Scale
                 & \textbf{45.0}/\textbf{62.5}/66.6
                 & \textbf{65.0}/\textbf{49.6}
                 & \textbf{53.7}/\textbf{53.9}/\textbf{53.5}                                                              \\
                \bottomrule
            \end{tabular}
            \label{tab:maskmem_module}
        }
    \end{minipage}\hfill
    \begin{minipage}[t]{0.5\textwidth}
        \centering
        \setlength{\tabcolsep}{2pt}
        \caption{Ablation on the memory size.}
        \resizebox{\textwidth}{!}{
            \begin{tabular}{ccccc}
                \toprule
                \multirow{3}{*}{\textbf{\#Mem. Size}}
                & \textbf{V-Gen.}          & \textbf{V-OV}  & \textbf{V-Ref.}                & \textbf{V-Rea.}           \\
                & \scriptsize Pan./Sem./Ins.        & \scriptsize Ins.        & \scriptsize YT21/DV17          & \scriptsize Ref./Rea./All \\
                & \scriptsize VPQ1/mIoU/mAP        & \scriptsize mAP         & \scriptsize \mjf               & \scriptsize \mjf          \\
                \midrule
                1
                & \underline{43.9}/62.0/\underline{68.0} & 58.5                    & \textbf{66.8}/\underline{50.7} & 56.9/56.9/56.5            \\
                2
                & \underline{43.9}/62.0/67.9             & 58.2                    & 66.5/\textbf{50.8}             & 57.6/57.7/57.2            \\
                4
                & 43.8/\textbf{63.2}/\textbf{68.4} & \underline{58.9} & \underline{66.7}/50.3         & \underline{57.7}/\textbf{58.0}/\underline{57.4} \\
                6
                & \underline{43.9}/\underline{62.6}/\textbf{68.4} & \textbf{60.2}    & 66.5/50.5                      & \textbf{58.0}/\underline{57.9}/\textbf{57.5}    \\
                8
                & \textbf{45.0}/62.5/66.6
                & 58.3
                & 65.0/49.6
                & 53.7/53.9/53.5                                                              \\
                \bottomrule
            \end{tabular}
            \label{tab:maskmem_size}
        }
    \end{minipage}
\end{table*}
 
\noindent\headbf{Mask Decoder.} We ablate the impact of the Token-to-Image (T2I) attention module and its initialization strategies in \tablename~\ref{tab:maskdec}. Compared to the baseline, employing a random initialization (+ T2I(Rand)) disrupts early training, occasionally yielding sub-optimal results and falling behind the baseline in static image tasks (e.g., I-Ref RefCOCO/+/g decreases from 82.9/78.0/79.5 to 82.5/77.2/79.4). Conversely, our proposed zero-initialization strategy (+ T2I(Zero)) facilitates a stable injection of the LLM's semantic token embeddings into the spatial features. This approach consistently outperforms the baseline, achieving peak performance across image metrics (e.g., 83.3/77.8/79.5 in I-Ref) and substantial gains in video tasks (e.g., increasing V-Ref YT21 from 53.6 to 60.8 \mjf).
\begin{table*}[t]
    \centering
    \setlength{\tabcolsep}{2pt}
    \caption{Comparison of state-of-the-art segmentation methods across image and video segmentation benchmarks, ranging from non-MLLM-based to MLLM-based, and from specialists to generalists. ``{\xmark}'' denotes unsupported. ``--'' indicates unreported. Best results are in \textbf{bold}, second-best are \underline{underlined}.}
    \resizebox{\textwidth}{!}{
        \begin{tabular}{lccccccc}
            \toprule
            \multirow{5}{*}{\textbf{Method}}
                                                                  & \multicolumn{5}{c}{\textbf{Textual Prompt}}
                                                                  & \multicolumn{2}{c}{\textbf{Visual Prompt}}                                                                                                                                                                                                                                                                                                     \\
            \cmidrule(lr){2-6} \cmidrule(lr){7-8}
                                                                  & \multicolumn{7}{c}{\textbf{\textit{Image Segmentation}}}                                                                                                                                                                                                                                                                                      \\
            \cmidrule(lr){2-8}
                                                                  & \textbf{\textit{I-Gen.}}                                 & \textbf{\textit{I-OV}}                             & \textbf{\textit{I-Ref.}}                           & \textbf{\textit{I-Rea.}}                        & \textbf{\textit{I-GCG}}           & \textbf{\textit{I-Int.}}       & \textbf{\textit{I-VGD}}           \\
                                                                  & \textit{Pan./Sem./Ins.}                                  & \textit{Pan./Sem./Ins.}                            & \textit{RefCOCO/+/g}                               & \textit{Val/Test}                               & \textit{Val/Test}                 & \textit{Point/Box}             & \textit{Point/Box}                \\
                                                                  & \scriptsize PQ/mIoU/mAP                                  & \scriptsize PQ/mIoU/mAP                            & \scriptsize cIoU                                   & \scriptsize gIoU                                & \scriptsize mIoU                  & \scriptsize mIoU               & \scriptsize mAP                   \\
            \midrule
            \rowcolor{LightGray} \multicolumn{8}{l}{\textbf{\textit{Non-MLLM-based Image Specialists}}}                                                                                                                                                                                                                                                                                           \\
            \color{gray} SAM-L~\cite{kirillov2023sam}             & \color{gray} \xmark                                      & \color{gray} \xmark                                & \color{gray} \xmark                                & \color{gray} \xmark                             & \color{gray} \xmark               & \color{gray} 51.8/76.6         & \color{gray} 12.8/31.7            \\
            \color{gray} Mask2Former-L~\cite{cheng2022maskformer} & \color{gray} 57.8/67.4/48.6                              & \color{gray} \xmark                                & \color{gray} \xmark                                & \color{gray} \xmark                             & \color{gray} \xmark               & \color{gray} \xmark            & \color{gray} \xmark               \\
            \color{gray} ODISE~\cite{xu2023odise}                 & \color{gray} 55.4/65.2/46.0                              & \color{gray} 22.6/29.9/14.4                        & \color{gray} \xmark                                & \color{gray} \xmark                             & \color{gray} \xmark               & \color{gray} \xmark            & \color{gray} \xmark               \\
            \midrule
            \rowcolor{LightGray} \multicolumn{8}{l}{\textbf{\textit{MLLM-based Image Generalists}}}                                                                                                                                                                                                                                                                                               \\
            LISA-7B~\cite{lai2024lisa}                            & \xmark                                                   & \xmark                                             & 74.9/65.1/67.9                                     & 52.9/47.3                           & \xmark                            & \xmark                         & \xmark                            \\
            GLaMM~\cite{rasheed2024glamm}                         & \xmark                                                   & \xmark                                             & 79.5/72.6/74.2                                     & \xmark                                          & 65.8/64.6             & \xmark                         & \xmark                            \\
            OMG-LLaVA~\cite{zhang2024omgllava}                    & 53.8/--/--                                   & \xmark                                             & 78.0/69.1/72.9                                     & \xmark                                          & 65.5/64.7             & \xmark                         & \xmark                            \\
            Sa2VA-8B~\cite{yuan2025sa2va}                         & \xmark                                                   & \xmark                                             & 81.6/76.2/78.7                         & --                                              & --                                & \xmark                         & \xmark                            \\
            X-SAM~\cite{wang2026xsam}                             & \textbf{54.7}/\textbf{66.5}/\textbf{47.0}                & \underline{20.9}/\underline{28.8}/\underline{16.2} & \textbf{85.1}/\underline{78.0}/\textbf{83.8}       & \underline{56.6}/\underline{57.8}               & \textbf{69.4}/\textbf{69.0}       & \underline{65.4}/\underline{69.6} & \textbf{47.9}/\textbf{49.5}     \\
            \rowcolor{LightGreen} \textbf{X2SAM}                   & \underline{54.1}/\underline{64.8}/\underline{45.8}                   & \textbf{31.2}/\textbf{38.2}/\textbf{20.2}          & \underline{84.0}/\textbf{78.4}/\underline{81.9}    & \textbf{71.1}/\textbf{68.5}                     & \underline{67.1}/\underline{65.2} & \textbf{67.7}/\textbf{70.3}    & \underline{45.9}/\underline{48.5} \\
            \midrule
            \multirow{5}{*}{\textbf{Method}}                      & \multicolumn{7}{c}{\textbf{\textit{Video Segmentation}}}                                                                                                                                                                                                                                                                                      \\
            \cmidrule(lr){2-8}
                                                                  & \textbf{\textit{V-Gen.}}                                 & \textbf{\textit{V-OV}}                             & \textbf{\textit{V-Ref.}}                           & \textbf{\textit{V-Rea.}}                        & \textbf{\textit{V-GCG}}           & \textbf{\textit{V-Obj.}}       & \textbf{\textit{V-VGD}}           \\
                                                                  & \textit{Pan./Sem./Ins.}                                  & \textit{YT21-Ins.}                                 & \textit{YT21/DV17}                                 & \textit{Ref./Rea./All}                          & \textit{V-GLaMM}                  & \textit{YT19-All}              & \textit{YT19/VIPSeg}                \\
                                                                  & \scriptsize PQ/mIoU/mAP                                  & \scriptsize mAP                                    & \scriptsize \mjf                                   & \scriptsize \mjf                                & \scriptsize mIoU                  & \scriptsize \mjf               & \scriptsize mAP                   \\
            \midrule
            \rowcolor{LightGray} \multicolumn{8}{l}{\textbf{\textit{Non-MLLM-based Video Specialists}}}                                                                                                                                                                                                                                                                                           \\
            \color{gray} SAM2-H~\cite{ravi2024sam2}               & \color{gray} \xmark                                      & \color{gray} \xmark                                & \color{gray} \xmark                                & \color{gray} \xmark                             & \color{gray} \xmark               & \color{gray} 88.8                & \color{gray} 39.2/25.6            \\
            \color{gray} OMG-Seg~\cite{li2024omgseg}              & \color{gray} 49.8/--/56.4                                & \color{gray} 50.5                                  & \color{gray} \xmark                                & \color{gray} \xmark                             & \color{gray} \xmark               & \color{gray} \xmark            & \color{gray} \xmark               \\
            \color{gray} UniRef++-L~\cite{wu2023uniref++}         & \color{gray} \xmark                                      & \color{gray} \xmark                                & \color{gray} 66.9/67.2                             & \color{gray} \xmark                             & \color{gray} \xmark               & \color{gray} 85.9              & \color{gray} \xmark               \\
            \midrule
            \rowcolor{LightGray} \multicolumn{8}{l}{\textbf{\textit{MLLM-based Video Generalists}}}                                                                                                                                                                                                                                                                                               \\
            VISA-7B~\cite{yan2024visa}                            & \xmark                                                   & \xmark                                             & 61.5/69.4                              & 50.9/43.0/46.9                                  & \xmark                            & \xmark                         & \xmark                            \\
VideoLISA~\cite{bai2024videolisa}                     & \xmark                                                   & \xmark                                             & 61.7/67.7                                          & --                                              & \xmark                            & \xmark                         & \xmark                            \\
UniPixel-7B~\cite{liu2025unipixel}                    & \xmark                                                   & \xmark                      & \underline{71.0}/\underline{76.4}                & 65.8/61.5/63.7           & \xmark                  & \xmark                   & \xmark                  \\
            HyperSeg~\cite{wei2024hyperseg}                       & --                                                       & --                                                 & 68.5/71.2                              & 58.5/53.0/55.7                                  & \xmark                            & \xmark                         & \xmark                            \\
            VideoGLaMM~\cite{munasinghe2024videoglamm}            & \xmark                                                   & \xmark                                             & 66.8/--                                            & --                                              & \underline{54.3}                  & \xmark                         & \xmark                            \\
            \rowcolor{LightGreen} \textbf{X2SAM}                   & \textbf{47.3}/\textbf{65.1}/\textbf{69.9}                & \textbf{60.3}                                      & \textbf{78.5}/\textbf{79.0}                                 & \textbf{69.3}/\textbf{70.7}/\textbf{69.9}       & \textbf{75.8}                     & \textbf{74.0}                  & \textbf{73.8}/\textbf{55.4}       \\
            \bottomrule
        \end{tabular}
    }
    \label{tab:performance}

\end{table*} \noindent\headbf{Joint Training.} \tablename~\ref{tab:training} evaluates the efficacy of our unified joint training strategy against separate and simple joint training paradigms. The simple joint training method achieves strong performance (e.g., 54.4 I-Gen.\ PQ and 64.6 V-Gen.\ mIoU) but requires approximately 5.2K GPU hours. Our unified joint training strategy reduces the training cost to 3.3K GPU hours, a 36.5\% reduction, while maintaining comparable image performance and improving video-level metrics such as V-Gen.\ mIoU (64.7) and V-OV Ins.\ mAP (59.1).

\noindent\headbf{Mask Memory.} As shown in \tablename~\ref{tab:maskmem_module}, directly adopting single-scale memory features provides only marginal improvements on V-Rea.\ and slightly degrades V-Gen.\ and V-Ref., indicating that naive temporal memory is insufficient. In contrast, introducing mask guidance brings consistent gains across tasks, especially improving V-Ref.\ YT21 from 53.6 to 63.3 \mjf, which demonstrates the importance of mask-level cues for temporal alignment. Class guidance further improves semantic discrimination, leading to stronger results on V-Rea. Finally, the full multi-scale design achieves the best overall performance, obtaining 45.0 VPQ, 62.5 mIoU, 65.0 \mjf\ on V-Ref.\ \textit{YT21}, and 53.5 \mjf\ on V-Rea.\ \textit{All}.

\noindent\headbf{Memory Size.} As reported in \tablename~\ref{tab:maskmem_size}, increasing the memory size from 1 to 6 generally benefits tasks requiring long-term temporal context, particularly open-vocabulary and reasoning-oriented video understanding. The best V-OV mAP of 60.2 and V-Rea.\ All score of 57.5 \mjf\ are achieved with memory size 6, suggesting that richer historical information helps resolve temporal ambiguity. However, the gains are not monotonic across all benchmarks. Some V-Gen.\ and V-Ref.\ metrics saturate earlier, and increasing the memory size to 8 leads to lower V-Ref.\ and V-Rea.\ performance despite improving V-Gen.\ VPQ. This indicates that excessive memory may introduce redundant or noisy temporal cues, while a moderate memory size provides a better trade-off, we adopt 6 frames as the final memory size.
\subsection{Benchmark Results}
\begin{table*}[t]
    \centering
    \setlength{\tabcolsep}{2pt}
    \caption{Comparison across image and video reasoning segmentation benchmarks.}
    \resizebox{\textwidth}{!}{
        \begin{tabular}{lccccccc}
            \toprule
            \multirow{3}{*}{\textbf{Method}}
                                                                & \multicolumn{4}{c}{\textbf{I-Rea. Seg.}}
                                                                & \multicolumn{3}{c}{\textbf{V-Rea. Seg.}}                                                                                                                                                                                                                                                                                                                                    \\
            \cmidrule(lr){2-5} \cmidrule(lr){6-8}
                                                                & \textbf{\textit{Val}}$_\textit{\textrm{Overall}}$ & \textbf{\textit{Test}}$_\textit{\textrm{Short}}$ & \textbf{\textit{Test}}$_\textit{\textrm{Long}}$ & \textbf{\textit{Test}}$_\textit{\textrm{Overall}}$ & \textbf{\textit{ReVOS}}$_\textit{\textrm{Ref.}}$   & \textbf{\textit{ReVOS}}$_\textit{\textrm{Rea.}}$   & \textbf{\textit{ReVOS}}$_\textit{\textrm{Overall}}$ \\
                                                                & \scriptsize cIoU/gIoU                             & \scriptsize cIoU/gIoU                            & \scriptsize cIoU/gIoU                           & \scriptsize cIoU/gIoU                              & \scriptsize \mj/\mf/\mjf                           & \scriptsize \mj/\mf/\mjf                           & \scriptsize \mj/\mf/\mjf                            \\
            \midrule
            \rowcolor{LightGray} \multicolumn{8}{l}{\textbf{\textit{Non-MLLM-Based Specialists}}}                                                                                                                                                                                                                                                                                                                                             \\
            \color{gray} SEEM-L~\cite{zou2023seem}              & \color{gray} 21.2/25.5                            & \color{gray} 11.5/20.1                           & \color{gray} 20.8/25.6                          & \color{gray} 18.7/24.4                             & \color{gray} --                                    & \color{gray} --                                    & \color{gray} --                                     \\
            \color{gray} ReferFormer-B~\cite{wu2022referformer} & \color{gray} --                                   & \color{gray} --                                  & \color{gray} --                                 & \color{gray} --                                    & \color{gray} 31.2/34.3/32.7                        & \color{gray} 21.3/25.6/23.4                        & \color{gray} 26.2/29.9/28.1                         \\
            \midrule
            \rowcolor{LightGray} \multicolumn{8}{l}{\textbf{\textit{MLLM-Based Generalists}}}                                                                                                                                                                                                                                                                                                                                                 \\
            LISA-7B~\cite{lai2024lisa}                          & 54.0/52.9                                         & \underline{40.6}/\underline{40.6}                & \underline{51.0}/\underline{49.4}               & \underline{48.4}/\underline{47.3}                  & 44.3/47.1/45.7                                     & 33.8/38.4/36.1                                     & 39.1/42.7/40.9                                      \\
            VISA-7B~\cite{yan2024visa}                          & \underline{57.8}/52.7                             & --                                               & --                                              & --                                                 & 49.2/52.6/50.9                                     & 40.6/45.4/43.0                                     & 44.9/49.0/46.9                                      \\
            HyperSeg~\cite{wei2024hyperseg}                     & 56.7/\underline{59.2}                             & --                                               & --                                              & --                                                 & \underline{56.0}/\underline{60.9}/\underline{58.5} & \underline{50.2}/\underline{55.8}/\underline{53.0} & \underline{53.1}/\underline{58.4}/\underline{55.7}  \\
            \rowcolor{LightGreen} \textbf{X2SAM}                 & \textbf{64.5}/\textbf{71.1}                       & \textbf{53.5}/\textbf{60.0}                      & \textbf{66.7}/\textbf{68.9}                     & \textbf{65.6}/\textbf{68.5}                        & \textbf{66.2}/\textbf{72.4}/\textbf{69.3}          & \textbf{67.5}/\textbf{74.0}/\textbf{70.3}          & \textbf{66.7}/\textbf{73.0}/\textbf{69.9}           \\
            \bottomrule
        \end{tabular}
    }
    \label{tab:rea-seg}
\end{table*}

\begin{table*}[t]
    \centering
    \setlength{\tabcolsep}{2pt}
    \caption{Comparison on out-of-domain tasks, including image generalized referring segmentation, image and video open-vocabulary segmentation benchmarks.}
\begin{tabular}{lccccccc}
            \toprule
            \multirow{3}{*}{\textbf{Method}}
             & \multicolumn{3}{c}{\textbf{I-Ref. Seg.}}
             & \multicolumn{3}{c}{\textbf{I-OV Seg.}}
             & \multicolumn{1}{c}{\textbf{V-OV Seg.}}                                             \\
            \cmidrule(lr){2-4} \cmidrule(lr){5-7} \cmidrule(lr){8-8}
             & \textbf{\textit{gRefCOCO}}$_\textit{\textrm{Val}}$
             & \textbf{\textit{gRefCOCO}}$_\textit{\textrm{TestA}}$
             & \textbf{\textit{gRefCOCO}}$_\textit{\textrm{TestB}}$
             & \textbf{\textit{A150}}$_\textit{\textrm{Pan.}}$
             & \textbf{\textit{A150}}$_\textit{\textrm{Sem.}}$
             & \textbf{\textit{A150}}$_\textit{\textrm{Ins.}}$
             & \textbf{\textit{YT-VIS-21}}$_\textit{\textrm{Ins.}}$                               \\
             & \scriptsize cIoU/gIoU
             & \scriptsize cIoU/gIoU
             & \scriptsize cIoU/gIoU
             & \scriptsize PQ
             & \scriptsize mIoU
             & \scriptsize mAP
             & \scriptsize AP/AP50                                                                \\
            \midrule
            \rowcolor{LightGray} \multicolumn{8}{l}{\textbf{\textit{Non-MLLM-Based Specialists}}} \\
            \color{gray} ReLA~\cite{liu2023gres}
             & \color{gray} 62.4/63.6
             & \color{gray} 69.3/70.0
             & \color{gray} 59.9/61.0
             & \color{gray} \xmark
             & \color{gray} \xmark
             & \color{gray} \xmark
             & \color{gray} \xmark                                                                \\
            \color{gray} ODISE~\cite{xu2023odise}
             & \color{gray} \xmark
             & \color{gray} \xmark
             & \color{gray} \xmark
             & \color{gray} 22.6
             & \color{gray} 29.9
             & \color{gray} 14.4
             & \color{gray} \xmark                                                                \\
            \color{gray} OMG-Seg~\cite{li2024omgseg}
             & \color{gray} \xmark
             & \color{gray} \xmark
             & \color{gray} \xmark
             & \color{gray} 27.9
             & \color{gray} --
             & \color{gray} --
             & \color{gray} 50.5/--                                                               \\
            \midrule
            \rowcolor{LightGray} \multicolumn{8}{l}{\textbf{\textit{MLLM-Based Generalists}}}     \\
            LISA-7B~\cite{lai2024lisa}
             & 38.7/--
             & 52.6/--
             & 44.8/--
             & \xmark
             & \xmark
             & \xmark
             & \xmark                                                                             \\
            PSALM~\cite{zhang2024psalm}
             & 42.0/\underline{43.3}
             & 52.4/\underline{54.5}
             & 50.6/\underline{52.5}
             & 13.7
             & 18.2
             & 9.0
             & \xmark                                                                             \\
            HyperSeg~\cite{wei2024hyperseg}
             & \underline{47.5}/--
             & \underline{57.3}/--
             & \underline{52.5}/--
             & 16.1
             & 22.3
             & --
             & \underline{53.8}/--                                                                \\
            X-SAM~\cite{wang2026xsam}
             & 59.9/65.6
             & 63.0/67.2
             & 62.0/65.2
             & \underline{20.9}
             & \underline{28.8}
             & \underline{16.2}
             & \xmark                                                                             \\
            \rowcolor{LightGreen} \textbf{X2SAM}
             & \textbf{63.1}/\textbf{68.1}
             & \textbf{67.3}/\textbf{71.2}
             & \textbf{63.4}/\textbf{66.7}
             & \textbf{31.2}
             & \textbf{38.2}
             & \textbf{20.2}
             & \textbf{60.3}/\textbf{78.0}                                                        \\
            \bottomrule
        \end{tabular}
\label{tab:ood-seg}
\end{table*}

\begin{table*}[tp]
    \centering
    \setlength{\tabcolsep}{2pt}
    \caption{Comparison across image and video visual grounded segmentation benchmarks.}
\begin{tabular}{lcccccc}
            \toprule
            \multirow{3}{*}{\textbf{Method}}
                                                & \multicolumn{2}{c}{\textbf{I-VGD Seg.}}
                                                & \multicolumn{4}{c}{\textbf{V-VGD Seg.}}                                                                                                                                                                                                                                                                               \\
            \cmidrule(lr){2-3} \cmidrule(lr){4-7}
                                                & \textbf{\textit{COCO}}$_\textit{\textrm{Point}}$ & \textbf{\textit{COCO}}$_\textit{\textrm{Box}}$ & \textbf{\textit{YT-VIS19}}$_\textit{\textrm{Point}}$ & \textbf{\textit{YT-VIS19}}$_\textit{\textrm{Box}}$ & \textbf{\textit{VIPSeg}}$_\textit{\textrm{Point}}$ & \textbf{\textit{VIPSeg}}$_\textit{\textrm{Box}}$ \\
            \cmidrule(lr){2-3} \cmidrule(lr){4-5} \cmidrule(lr){6-7}
                                                & \scriptsize AP/AP50                              & \scriptsize AP/AP50                            & \scriptsize AP/AP50                                  & \scriptsize AP/AP50                                & \scriptsize AP/AP50                                & \scriptsize AP/AP50                              \\
            \midrule
            \rowcolor{LightGray} \multicolumn{7}{l}{\textbf{\textit{Non-MLLM-Based Specialists}}}                                                                                                                                                                                                                                                                       \\
            \color{gray} SAM-L~\cite{kirillov2023sam}                  & \color{gray} 12.8/22.8                           & \color{gray} 31.7/50.1                         & \color{gray} --                                      & \color{gray} --                                    & \color{gray} --                                    & \color{gray} --                                  \\
            \color{gray} SAM2-H~\cite{ravi2024sam2}                 & \color{gray} --                                  & \color{gray} --                                & \color{gray} 39.2/53.5                               & \color{gray} 54.0/73.3                             & \color{gray} 25.6/36.3                             & \color{gray} 40.4/54.7                           \\
            \midrule
            \rowcolor{LightGray} \multicolumn{7}{l}{\textbf{\textit{MLLM-Based Specialists}}}                                                                                                                                                                                                                                                                           \\
            PSALM~\cite{zhang2024psalm}         & 2.0/3.3                                          & 3.7/5.8                                        & \xmark                                               & \xmark                                             & \xmark                                             & \xmark                                           \\
            X-SAM~\cite{wang2026xsam}           & \textbf{47.9/72.5}                               & \textbf{49.5/74.7}                             & \xmark                                               & \xmark                                             & \xmark                                             & \xmark                                           \\
            \rowcolor{LightGreen} \textbf{X2SAM} & \underline{45.9}/\underline{68.2}                            & \underline{48.5}/\underline{71.6}                          & \textbf{73.8}/\textbf{93.5}                                   & \textbf{74.4}/\textbf{93.9}                                 & \textbf{55.5}/\textbf{75.4}                                 & \textbf{57.8}/\textbf{78.3}                               \\
            \bottomrule
        \end{tabular}
\label{tab:vgd-seg}
\end{table*}
 
\noindent\headbf{Overall Performance.}
\tablename~\ref{tab:performance} presents a comparison of X2SAM against specialist models and MLLM-based generalists across image and video segmentation benchmarks. On image tasks, X2SAM remains competitive with the image-centric generalist X-SAM~\cite{wang2026xsam} while improving image open-vocabulary segmentation (I-OV) from 20.9 to 31.2 PQ. It also maintains strong performance on generic (I-Gen.) and referring (I-Ref.) segmentation, and obtains 67.1/65.2 mIoU on I-GCG validation/test splits. These results suggest that extending the architecture to video does not collapse its static image segmentation ability.
On video tasks, X2SAM outperforms existing MLLM-based video generalists on most reported video segmentation benchmarks. Compared with UniPixel-7B~\cite{liu2025unipixel}, X2SAM improves V-Ref.\ on both Ref-YT21 and Ref-DV17, and achieves strong V-Gen.\ performance with 65.1 mIoU. On video grounded conversation generation (V-GCG), X2SAM improves over VideoGLaMM~\cite{munasinghe2024videoglamm} by +21.5 mIoU (75.8 vs. 54.3). These gains support the role of the unified formulation and mask memory in video-level segmentation, while the remaining gap to task-specific video object segmentation (VOS) specialists highlights the cost of maintaining a general interface.

\noindent\headbf{Reasoning Segmentation.}
\tablename~\ref{tab:rea-seg} reports X2SAM's performance on reasoning segmentation, where targets require implicit knowledge and logical deduction. On I-Rea.\ Seg., X2SAM achieves 64.5 cIoU and 71.1 gIoU on validation, outperforming HyperSeg~\cite{wei2024hyperseg} by +7.8 cIoU and +11.9 gIoU. It also sets new state-of-the-art test results for both short and long queries, showing robust instruction understanding. On V-Rea.\ Seg., X2SAM reaches 69.9 \mjf on ReVOS, improving over HyperSeg~\cite{wei2024hyperseg} by +14.2 points and surpassing the video-specialist ReferFormer-B~\cite{wu2022referformer}. Its consistent performance on both Ref. and Rea. subsets suggests effective integration of LLM-based target reasoning and spatio-temporal mask tracking.

\noindent\headbf{Out-of-Domain Segmentation.}
As shown in \tablename~\ref{tab:ood-seg}, we evaluate X2SAM on out-of-domain tasks covering unseen datasets and novel categories. On gRefCOCO~\cite{liu2023gres}, which includes multi-target and no-target expressions, X2SAM improves over both the non-MLLM specialist ReLA~\cite{liu2023gres} and recent MLLM-based generalists such as PSALM~\cite{zhang2024psalm} and X-SAM~\cite{wang2026xsam}. On ADE20K (A150)~\cite{zhou2019ade20k}, X2SAM achieves $31.2$ PQ, $38.2$ mIoU, and $20.2$ mAP, outperforming compared open-vocabulary segmentation methods including ODISE~\cite{xu2023odise} and X-SAM. In the video domain, X2SAM obtains $60.3$ AP on YT-VIS-21, exceeding OMG-Seg~\cite{li2024omgseg} and HyperSeg~\cite{wei2024hyperseg}. These results indicate that the unified training setup transfers beyond the primary training distribution, although the evaluation remains bounded by the available OOD benchmarks.

\noindent\headbf{Visual Grounded Segmentation.}
\tablename~\ref{tab:vgd-seg} presents a comparative analysis of visual grounded segmentation performance utilizing point and bounding box prompts. Within the image domain (I-VGD Seg.), X2SAM demonstrates highly competitive efficacy on the COCO benchmark, achieving an AP of 45.9 and 48.5 for point and box prompts, respectively, thereby performing comparably to the image-specialized X-SAM model. Crucially, the temporal modeling advantages of our approach become distinctly evident in the video domain (V-VGD Seg.), where X2SAM substantially improves over SAM2-H under the evaluated V-VGD protocol. Specifically, under box prompt evaluation, X2SAM attains an impressive 74.4 AP on YT-VIS19 and 57.8 AP on the more complex VIPSeg dataset, exhibiting significant improvements over SAM2-H (54.0 AP and 40.4 AP). These substantial performance gains validate the effectiveness of our region sampler and mask memory modules in robustly propagating temporal visual prompts across highly dynamic video frames.

\section{Discussion}
\noindent\headbf{Conclusion.}
We presented X2SAM, a unified segmentation-oriented MLLM for pixel-level understanding across images and videos. X2SAM casts diverse segmentation tasks into a shared instruction-following formulation, supporting both textual instructions and visual prompts within the same framework. By integrating a Mask Memory module, it extends image-centric any-segmentation capabilities to video sequences with improved temporal consistency. We also introduced the Video Visual Grounded (V-VGD) benchmark to evaluate video object grounding from interactive visual prompts. In addition, our adaptive joint training strategy enables efficient co-training over heterogeneous image and video datasets, reducing the training cost while maintaining balanced performance across modalities. Extensive experiments show that X2SAM achieves a strong balance between task coverage and accuracy: it remains competitive on image segmentation benchmarks, improves many video segmentation tasks, and preserves general image and video understanding abilities.

\noindent\headbf{Limitations and Outlook.}
X2SAM still has several limitations. First, unified training over heterogeneous image and video datasets remains computationally expensive, especially for video samples with high memory cost. Second, the fixed-size FIFO memory may be insufficient for long videos with prolonged occlusions, large appearance changes, or sparse target reappearance. Third, as a unified generalist model, X2SAM may still lag behind specialized models on narrowly focused tasks such as optimized video object segmentation or image-only segmentation. Future work will explore more efficient training, lightweight backbones, and adaptive long-range memory to improve scalability and robustness.

\newpage
\appendix
\section{More Formulation Details}
\label{sec:app_formulation}
\begin{table*}[!ht]
    \centering
    \setlength{\tabcolsep}{2pt}
    \caption{Prompt templates for image and video segmentation tasks.}
    \label{tab:segmentation}
    \resizebox{\textwidth}{!}{
        \begin{tabular}{>{\centering\arraybackslash}m{2cm}>{\centering\arraybackslash}m{1.2cm}>{\raggedright\arraybackslash}m{5.5cm}>{\centering\arraybackslash}m{1.2cm}>{\raggedright\arraybackslash}m{5.5cm}}
            \toprule
            \multirow{2}{*}{\textbf{Prompt}}
             & \multicolumn{2}{c}{\textbf{Image Segmentation}}
             & \multicolumn{2}{c}{\textbf{Video Segmentation}}                                                                                                                                                                                                                                                   \\
            \cmidrule(lr){2-3} \cmidrule(lr){4-5}
             & \makecell{\textbf{Task}}                                                                                                                                                                                   & \makecell{\textbf{Example}} & \makecell{\textbf{Task}} & \makecell{\textbf{Example}} \\
            \midrule
            person, hat, tree\ldots
             & I-Gen.
             & Can you segment the image based on the following categories: \texttt{<p>person</p>}, \texttt{<p>person</p>}, \texttt{<p>tree</p>}, \ldots? Please output the segmentation masks.
             & V-Gen.
             & Can you provide segmentation masks for this video based on these categories: \texttt{<p>person</p>}, \texttt{<p>person</p>}, \texttt{<p>tree</p>}, \ldots? Please provide the segmentation masks.                                                                                                 \\
            \midrule
            phone, box, human\ldots
             & I-OV
             & Can you provide segmentation masks for this image based on these categories: \texttt{<p>phone</p>}, \texttt{<p>box</p>}, \texttt{<p>human</p>}, \ldots? Please provide the segmentation masks.
             & V-OV
             & Could you create segmentation masks for this video according to the specified categories: \texttt{<p>phone</p>}, \texttt{<p>box</p>}, \texttt{<p>human</p>}, \ldots? Please create the segmentation masks.                                                                                        \\
            \midrule
            the right man
             & I-Ref.
             & Please identify and segment the \texttt{<p>the right man</p>} in this image.
             & V-Ref.
             & What is \texttt{<p>the right man</p>} in this video? Please output the corresponding segmentation mask.                                                                                                                                                                                           \\
            \midrule
            What can be used to contact others?
             & I-Rea.
             & \texttt{<p>What can be used to contact others in this image</p>}? Please segment the image.
             & V-Rea.
             & \texttt{<p>What can be used to contact others in this video</p>}? Please segment the video.                                                                                                                                                                                                       \\
            \midrule
            None
             & I-GCG
             & Can you provide a brief description of this image? Please output interleaved segmentation masks for the corresponding phrases.
             & V-GCG
             & Could you give me a brief explanation of this video? Please respond with interleaved segmentation masks for the corresponding phrases.                                                                                                                                                          \\
            \midrule
            V-Prompt
             & I-Int.
             & Can you segment the image based on the following regions: \texttt{<p><region></p>}? Please output the corresponding segmentation mask.
             & V-Obj.
             & Could you create segmentation masks for this video according to the specified regions: \texttt{<p><region></p>}? Please create the segmentation masks.                                                                                                                                            \\
            \midrule
            V-Prompt
             & I-VGD
             & Can you provide segmentation masks for this video based on these regions: \texttt{<p><region></p>}, \texttt{<p><region></p>}, \ldots? Please provide the segmentation masks.
             & V-VGD
             & Could you output segmentation masks for this video that highlight the following regions: \texttt{<p><region></p>}, \texttt{<p><region></p>}, \ldots? Please output the segmentation masks.                                                                                                        \\
            \bottomrule
        \end{tabular}
    }
\end{table*}

\noindent\headbf{Prompt Templates.}
\tablename~\ref{tab:segmentation} summarizes the task-specific prompt templates used in our unified formulation. Across all image and video segmentation tasks, the target category, referring expression, reasoning query, or region prompt is represented as a conditional state wrapped by the special tokens \texttt{<p>} and \texttt{</p>}. Correspondingly, the LLM generates a dedicated \texttt{<SEG>} token for each object of interest, and the hidden representation of this token is used as the mask-aware directive for the mask decoder. In this way, heterogeneous tasks can be consistently cast into a shared language-conditioned segmentation interface.

\noindent Specifically, generic segmentation and open-vocabulary segmentation instantiate the conditional states with category names, while referring and reasoning segmentation use free-form natural language descriptions or queries. GCG segmentation follows an interleaved generation format, where phrase-level textual responses are aligned with the corresponding segmentation outputs. For object-centric tasks, including image interactive segmentation and video object segmentation, as well as visual grounding tasks, the conditional states are constructed from visual prompts or region placeholders, enabling the model to ground user-specified regions in either images or videos. Although the wording of the templates varies across task types and modalities, they all preserve the same structural principle: explicitly mark the object condition in the textual prompt and align each condition with a segmentation output token, thereby facilitating unified joint training over diverse datasets.
\section{More Dataset Details}
\label{sec:app_dataset}
\begin{table}[t]
    \centering
    \setlength{\tabcolsep}{4pt}
    \caption{Datasets for X2SAM across image and video domains. \textcolor{gray}{Grayed} datasets are only for evaluation and not used in training.}
    \label{tab:datasets}
\begin{tabular}{lcr}
            \toprule
            \multicolumn{3}{c}{\textbf{Image Domain}}                                                                                                                                                    \\
            \textbf{Task} & \textbf{Datasets}                                                                                                                                        & \textbf{\#Data}   \\
            \midrule
            I-Chat        & LLaVA-1.5~\cite{liu2024llava1x5}, \textcolor{gray}{Image Benchmarks}~\cite{fu2024mme,liu2024mmbench,li2024seed,li2023pope,kembhavi2016ai2d}                                                                                                                        & 624.6K            \\
            I-Gen.        & COCO~\cite{lin2014coco}                                                                                                                                  & 118.3K            \\
            I-OV          & \textcolor{gray}{ADE20k}~\cite{zhou2019ade20k}                                                                                                           & /                 \\
            I-Ref.        & RefCOCO, RefCOCO+, RefCOCOg, \textcolor{gray}{gRefCOCO}~\cite{liu2023gres}                                                                               & 302.4K            \\
            I-Rea.        & ReasonSeg~\cite{lai2024lisa}                                                                                                                             & 0.2K              \\
            I-GCG         & Grand-f$_{\mathrm{GCG}}$, RefCOCOg$_{\mathrm{GCG}}$, PSG$_{\mathrm{GCG}}$, Fickr$_{\mathrm{GCG}}$~\cite{rasheed2024glamm}                                & 196.1K            \\
            I-Int.        & \textcolor{gray}{COCO-Int.}~\cite{zhang2024psalm}                                                                                                        & /                 \\
            I-VGD         & COCO-VGD~\cite{wang2026xsam}                                                                                                                             & 117.3K            \\
            \midrule
            \multicolumn{3}{c}{\textbf{Video Domain}}                                                                                                                                                    \\
            \textbf{Task} & \textbf{Datasets}                                                                                                                                        & \textbf{\#Sample} \\
            \midrule
            V-Chat        & VideoChatGPT~\cite{maaz2024videochatgpt}, \textcolor{gray}{Video Benchmarks}~\cite{fu2025videomme,li2024mvbench,zhou2025mlvu,wu2024longvideobench}                                                                                                                 & 13.3K             \\
            V-Gen.        & VIPSeg~\cite{miao2022vipseg}, VSPW~\cite{miao2021vspw}, YT-VIS19~\cite{yang2019ytvis}                                                                    & 30.7K             \\
            V-OV          & \textcolor{gray}{YT-VIS21}~\cite{yang2019ytvis}                                                                                                          & /                 \\
            V-Ref.        & YT-RefVOS21~\cite{seo2020urvos}, DAVIS17-RefVOS~\cite{perazzi2016davis}                                                                                  & 14.3K             \\
            V-Rea.        & ReVOS~\cite{yan2024visa}                                                                                                                                 & 18.4K             \\
            V-GCG         & MeVIS$_{\mathrm{GCG}}$, YT-VOS$_{\mathrm{GCG}}$, VidSTG$_{\mathrm{GCG}}$, HCSTV$_{\mathrm{GCG}}$, Video$_{\mathrm{GCG}}$~\cite{munasinghe2024videoglamm} & 107.9K            \\
            V-Obj.        & YT-VOS19~\cite{xu2018ytvos}                                                                                                                              & 13.4K             \\
            V-VGD         & YT19-VGD, VIPSeg-VGD                                                                                                                                     & 16.3K             \\
            \bottomrule
        \end{tabular}
\end{table} 

\noindent\headbf{Training Datasets.}
As summarized in \tablename~\ref{tab:datasets}, our training pipeline consists of two stages. We first train the class-agnostic segmentor on the mask-only SA-1B dataset~\cite{kirillov2023sam} to initialize the mask decoder. We then perform unified joint training over all image and video domains. For image domain tasks, we follow the mixed fine-tuning setup of X-SAM~\cite{wang2026xsam}, covering COCO for generic segmentation, RefCOCO/RefCOCO+/RefCOCOg for referring segmentation, ReasonSeg~\cite{lai2024lisa} for reasoning segmentation, several GCG datasets from GLaMM~\cite{rasheed2024glamm}, COCO-VGD for visual grounded segmentation, and LLaVA-1.5~\cite{liu2024llava1x5} for image chat. For video domain tasks, we use VIPSeg~\cite{miao2022vipseg}, VSPW~\cite{miao2021vspw}, and YT-VIS19~\cite{yang2019ytvis} for generic segmentation; YT-RefVOS21~\cite{seo2020urvos} and DAVIS17-RefVOS~\cite{perazzi2016davis} for referring segmentation; ReVOS~\cite{yan2024visa} for reasoning segmentation; VideoGLaMM-derived datasets for grounded conversation generation~\cite{munasinghe2024videoglamm}; YT-VOS19~\cite{xu2018ytvos} for video object segmentation; and our newly constructed YT19-VGD and VIPSeg-VGD datasets for video visual grounded segmentation. For video chat, we use the VideoInstruct100K corpus from VideoChatGPT~\cite{maaz2024videochatgpt}. The grayed entries in \tablename~\ref{tab:datasets} denote datasets that are reserved for evaluation and are not used during training.

\noindent\headbf{Evaluation Datasets.}
We evaluate X2SAM on the standard validation or test splits of each benchmark following prior work. In-domain evaluation covers all 14 image and video segmentation tasks listed in \tablename~\ref{tab:datasets}, using the corresponding benchmark protocols and metrics described in Section~\ref{sec:task_datas_metrics}. To further assess generalization, we additionally report out-of-domain performance on gRefCOCO~\cite{liu2023gres} for image referring segmentation, ADE20K~\cite{zhou2019ade20k} for image open-vocabulary segmentation, and YT-VIS-21~\cite{yang2019ytvis} for video open-vocabulary segmentation. These datasets are excluded from training and are used only for zero-shot or transfer evaluation.

\noindent\headbf{Video VGD Datasets Construction.}
We construct two video visual grounded segmentation datasets, namely YT19-VGD and VIPSeg-VGD, by extending the original annotations of YT-VIS19~\cite{yang2019ytvis} and VIPSeg~\cite{miao2022vipseg} into a unified visual-prompt-driven format. Similar to COCO-VGD, each target object is paired with four automatically generated visual prompts, including point, scribe, box, and mask, following~\cite{zhang2024psalm}. For each object trajectory, the prompt is generated from its first visible annotated frame and serves as the grounded condition, while the supervision target is the full spatio-temporal mask sequence of that object throughout the video clip. During training, we randomly sample one prompt type for each target instance to improve robustness to diverse visual grounding signals. During evaluation, we mainly report the point- and box-based settings, following the common protocol in visual prompt segmentation.
\noindent For YT19-VGD, we directly build upon the instance-level annotations of YT-VIS19, where each annotated object track naturally defines one video-grounded target. For VIPSeg-VGD, we derive the dataset from the panoptic annotations of VIPSeg and retain only \emph{thing} categories with valid instance identities across frames, so that each dynamic object can be converted into an instance trajectory for visual grounding. In this way, the two sub-datasets complement each other: YT19-VGD emphasizes instance-centric object videos in the wild, while VIPSeg-VGD introduces more challenging scene dynamics and denser multi-object contexts. Together, they form our V-VGD benchmark used for both unified training and evaluation.
\section{More Model Details}
\label{sec:app_model}

\noindent\headbf{Mask Encoder.} We implement the mask encoder following SAM2~\cite{ravi2024sam2}. Given an input image or video frame $X_v$, the mask encoder $g_m(\cdot)$ extracts dense mask-aware features $Z_m = g_m(X_v, M)$. The encoder is initialized from the pretrained SAM2 mask encoder, which provides fine-grained spatial representations for object boundaries, local textures, and region-level visual cues. Different from SAM2, we discard the original mask decoder and mask memory, and only retain the mask encoder as a lightweight feature extractor. The resulting mask features are projected into the multimodal latent space and used as fine-grained visual conditions for subsequent language-guided mask prediction.

\noindent\headbf{Mask Decoder.}
We follow X-SAM~\cite{wang2026xsam} to design the mask decoder, which is adapted from Mask2Former~\cite{cheng2022mask2former}. To incorporate language-conditioned information, we introduce a Token-to-Image Attention module that injects the latent embedding of the special \texttt{<SEG>} token into dense mask decoding. Specifically, the hidden state corresponding to \texttt{<SEG>} is transformed into mask queries $Q_m$, which attend to visual and memory-enhanced features $Y_m = g_\psi(Q_m, Z_w, Z_p)$, where $Z_w$ denotes the memory-augmented visual representation and $Z_p$ denotes the projected multimodal latent representation. This design enables the decoder to generate masks that are spatially accurate and semantically aligned with the language instruction.

\noindent\headbf{Mask Memory.}
Inspired by the memory architecture of SAM2~\cite{ravi2024sam2}, we design a mask memory module $g_\omega$ to store guided vision features from previous video frames. For frame $t$, the memory bank maintains the guided vision features of the most recent $K$ processed frames $\mathcal{B}_t = \{Z_{m}^{t-k}\}_{k=1}^{K}$, where the default capacity is $K=8$ for ablation studies while we set $K=6$ for the final training. The Memory Attention module attends to $\mathcal{B}_t$ and the current-frame visual representation to produce temporally-refined vision features $Z_w^t$ for the Mask Decoder. After the decoder predicts the current segmentation mask, the Memory Encoder encodes the downsampled vision features together with the current mask logits into a guided vision feature $Z_m^t$. The Memory Bank then stores $Z_m^t$ and updates the cache using a FIFO strategy. This fixed-size memory design improves temporal consistency for video segmentation while keeping the computational cost bounded.
\section{More Implementation Details}
\label{sec:app_implementation}
\begin{table}[t]
    \centering
    \caption{Hyper-parameters for generic segmentor and unified joint training.}
    \label{tab:parameters}
    \setlength{\tabcolsep}{2pt}
    \resizebox{\textwidth}{!}{
        \begin{tabular}{lcc}
            \toprule
            \textbf{Item}          & \textbf{Agnostic Segmentor Training}                 & \textbf{Unified Joint Training}                      \\
            \midrule
            gpu number             & 32                                                    & 32                                                    \\
            batch size of each device             & 1                                                    & 1                                                    \\
            batch multiplier(image/video) & 1/--                                                   & 4/1                                                    \\
            accumulative counts(image/video) & 1/--                                                   & 1/4                                                    \\
            training epochs        & 1                                                    & 1                                                    \\
            training modules       & mask decoder                                          & projectors, LLM, mask encoder \& decoder \& memory \\
            LoRA~\cite{hu2022lora} & --                                                   & $r=128$, $\alpha=256$                                \\
            max norm               & 0.01                                                 & 1                                                    \\
            lr for mask encoder    & --                                                   & \text{1e-5}                                          \\
            lr for other modules   & \multicolumn{2}{c}{ 1e-4}                                          \\
            lr schedule            & \multicolumn{2}{c}{Cosine Annealing}                                                                        \\
            optimizer              & \multicolumn{2}{c}{AdamW~\cite{loshchilov2017adamw}}                                                        \\
            optimizer momentum     & \multicolumn{2}{c}{$\beta_1=0.9$, $\beta_2=0.999$}                                                          \\
            warmup ratio           & \multicolumn{2}{c}{0.03}                                                                                    \\
            \bottomrule
        \end{tabular}
    }
\end{table} 

\noindent\headbf{Preprocessing.} To increase the number of training clips from video segmentation datasets, we adopt consecutive frame sampling with stride 1 and frame length 8 for most video segmentation datasets. For the video GCG segmentation task, we use a global sampling strategy and sample 16 frames for each video clip. For the video chat task, we sample 64 frames as the input to the vision encoder to support long-range video understanding.

\noindent\headbf{Evaluation.} We evaluate X2SAM on image and video segmentation benchmarks following the standard protocols of the corresponding benchmarks. For Video GCG segmentation, we report the average over the three sub-datasets shown in \tablename~\ref{tab:gcg-seg}. This protocol differs from the original VideoGLaMM~\cite{munasinghe2024videoglamm} reporting, so we compare against both reported and re-evaluated baselines whenever available.

\noindent\headbf{Training.} For agnostic segmentor training, we freeze the mask encoder and only optimize the mask decoder as a binary classification task with mask supervision. For unified joint training, we unfreeze the mask encoder and jointly optimize the mask encoder, projectors, LLM, mask decoder, and mask memory as a multi-task learning task with language generation and mask prediction.
\begin{table*}[!t]
    \centering
    \setlength{\tabcolsep}{2pt}
    \caption{Ablation on the data source of agnostic segmentor training.}
    \resizebox{\textwidth}{!}{
        \begin{tabular}{lcccccccc}
            \toprule
            \multirow{4}{*}{\makecell{\textbf{Data}\\\textbf{Source}}}
                                                                & \multicolumn{4}{c}{\textbf{Image Segmentation}}
                                                                & \multicolumn{4}{c}{\textbf{Video Segmentation}}                                                                                                                          \\
            \cmidrule(lr){2-5} \cmidrule(lr){6-9}
                                                                & \textbf{I-Gen.}                   & \textbf{I-OV}                      & \textbf{I-Ref.}          & \textbf{I-Rea.}   & \textbf{V-Gen.}                   & \textbf{V-OV} & \textbf{V-Ref.}    & \textbf{V-Rea.}          \\
                                                                & \textit{Pan./Sem./Ins.}        & \textit{Pan./Sem./Ins.}         & \textit{RefCOCO/+/g}  & \textit{Val/Test} & \textit{Pan./Sem./Ins.}      & \textit{Ins.} & \textit{YT21/DV17} & \textit{Ref./Rea./All} \\
                                                                & \scriptsize PQ/mIoU/mAP           & \scriptsize PQ/mIoU/mAP            & \scriptsize cIoU         & \scriptsize gIoU  & \scriptsize PQ/mIoU/mAP           & \scriptsize mAP & \scriptsize \mjf   & \scriptsize \mjf          \\
            \midrule
            COCO      & 50.9/63.1/41.1
                      & 27.4/34.9/16.1
                      & \textbf{82.2}/\underline{76.3}/\underline{79.3}
                      & 61.0/59.9
                      & \underline{43.7}/\underline{63.4}/\underline{69.4}
                      & \underline{59.2}
                      & \underline{68.3}/\underline{51.3}
                      & 58.0/58.5/57.8 \\
            SAM-Sub   & \underline{51.4}/\underline{63.3}/\underline{42.9}
                      & \underline{30.3}/\underline{36.6}/\underline{19.0}
                      & \underline{82.0}/76.1/79.1
                      & \underline{66.5}/\underline{65.3}
                      & 42.3/61.6/69.0
                      & 58.7
                      & \underline{68.3}/51.2
                      & \underline{58.6}/\underline{58.6}/\underline{58.2} \\
            SAM-1B    & \textbf{52.5}/\textbf{63.6}/\textbf{44.6}
                      & \textbf{31.4}/\textbf{38.1}/\textbf{20.2}
                      & \underline{82.0}/\textbf{76.8}/\textbf{79.6}
                      & \textbf{66.8}/\textbf{66.1}
                      & \textbf{45.4}/\textbf{63.5}/\textbf{69.8}
                      & \textbf{59.5}
                      & \textbf{68.5}/\textbf{51.5}
                      & \textbf{59.3}/\textbf{60.0}/\textbf{59.2} \\
            \bottomrule
        \end{tabular}
    }
    \label{tab:seg-comparison}
\end{table*}

\begin{table*}[!t]
    \centering
    \setlength{\tabcolsep}{2pt}
    \caption{Comparison of MLLMs on image and video segmentation benchmarks.}
    \resizebox{\textwidth}{!}{
        \begin{tabular}{lcccccccccc}
            \toprule
            \multirow{4}{*}{\textbf{MLLM}}
                & \multicolumn{5}{c}{\textbf{Image Segmentation}}
                & \multicolumn{5}{c}{\textbf{Video Segmentation}} \\
            \cmidrule(lr){2-6} \cmidrule(lr){7-11}
                & \textbf{I-Gen.} & \textbf{I-OV} & \textbf{I-Ref.} & \textbf{I-Rea.} & \textbf{I-VGD}
                & \textbf{V-Gen.} & \textbf{V-OV} & \textbf{V-Ref.} & \textbf{V-Rea.} & \textbf{V-VGD} \\
                & \textit{Pan./Sem./Ins.} & \textit{Pan./Sem./Ins.} & \textit{RefCOCO/+/g} & \textit{Val/Test} & \textit{Point/Box}
                & \textit{Pan./Sem./Ins.} & \textit{Ins.} & \textit{YT21/DV17} & \textit{Ref./Rea./All} & \textit{YT19/VIPSeg} \\
                & \scriptsize PQ/mIoU/mAP & \scriptsize PQ/mIoU/mAP & \scriptsize cIoU & \scriptsize gIoU & \scriptsize mAP
                & \scriptsize PQ/mIoU/mAP & \scriptsize mAP & \scriptsize \mjf & \scriptsize \mjf & \scriptsize mAP \\
            \midrule
            Siglip2+Phi3-3.8B
                & 53.8/64.7/46.0
                & 26.8/34.0/\underline{18.6}
                & 80.0/71.3/75.5
                & 61.9/60.0
                & 47.0/49.0
                & 46.1/\underline{64.0}/\textbf{70.7}
                & \textbf{63.7}
                & 66.2/50.2
                & 55.9/56.4/55.8
                & \textbf{73.6}/56.1 \\
            Siglip2+Qwen3-4B
                & \underline{54.2}/\textbf{65.3}/\underline{46.5}
                & \underline{27.9}/\underline{35.7}/17.8
                & \underline{81.0}/\underline{73.5}/\underline{77.9}
                & \underline{62.5}/\underline{61.8}
                & \underline{47.1}/\underline{49.5}
                & \underline{47.6}/\textbf{65.1}/\textbf{70.7}
                & \underline{62.2}
                & \underline{67.9}/\underline{51.5}
                & \underline{57.7}/\underline{58.3}/\underline{57.6}
                & \textbf{73.6}/\underline{56.4} \\
            Qwen3VL-4B
                & \textbf{54.7}/\underline{65.1}/\textbf{46.6}
                & \textbf{29.8}/\textbf{36.2}/\textbf{19.4}
                & \textbf{82.9}/\textbf{76.5}/\textbf{79.9}
                & \textbf{64.8}/\textbf{65.7}
                & \textbf{47.5}/\textbf{49.8}
                & \textbf{47.9}/\textbf{65.1}/\underline{70.0}
                & 61.2
                & \textbf{68.4}/\textbf{52.2}
                & \textbf{59.4}/\textbf{59.9}/\textbf{59.1}
                & \textbf{73.6}/\textbf{57.3} \\
            \bottomrule
        \end{tabular}
    }
    \label{tab:mllm}
\end{table*}

\begin{table}[!t]
    \centering
    \setlength{\tabcolsep}{4pt}
    \caption{Ablation on the region sampler, including sampler features and kernel size.}
    \resizebox{\textwidth}{!}{
    \begin{tabular}{cccccccccccc}
        \toprule
        \multicolumn{2}{c}{\multirow{2}{*}{\textbf{Sampler Features}}} & \multirow{2}{*}{\makecell{\textbf{Kernel}\\\textbf{Size}}}
            & \multicolumn{2}{c}{\textbf{I-Int}}
            & \multicolumn{2}{c}{\textbf{I-VGD}}
            & \multicolumn{4}{c}{\textbf{V-VGD}}
            & \textbf{V-Obj} \\
        \cmidrule(lr){4-5} \cmidrule(lr){6-7} \cmidrule(lr){8-11} \cmidrule(lr){12-12}
        & &
            & \textbf{\textit{Point}} & \textbf{\textit{Box}}
            & \textbf{\textit{Point}} & \textbf{\textit{Box}}
            & \multicolumn{2}{c}{\textbf{\textit{YT-VIS}}}
            & \multicolumn{2}{c}{\textbf{\textit{VIPSeg}}}
            & \textbf{\textit{Seen/Unseen/All}} \\
        \cmidrule(lr){1-2} \cmidrule(lr){3-3} \cmidrule(lr){4-5} \cmidrule(lr){6-7} \cmidrule(lr){8-9} \cmidrule(lr){10-11} \cmidrule(lr){12-12}
        \textbf{Vision Enc.} & \textbf{Mask Enc.} & \scriptsize K & \scriptsize mIoU & \scriptsize mIoU & \scriptsize AP & \scriptsize AP & \scriptsize AP$_{Point}$ & \scriptsize AP$_{Box}$ & \scriptsize AP$_{Point}$ & \scriptsize AP$_{Box}$ & \scriptsize \mjf \\
        \midrule
        \checkmark &            & $\infty$ & 64.7 & 70.7 & 44.2 & 46.7 & 72.1 & 73.2 & \textbf{50.6} & \textbf{54.8} & 65.6/57.2/64.6 \\
                   & \checkmark & $\infty$ & \textbf{66.2} & \textbf{72.1} & \textbf{45.2} & \textbf{47.1} & \textbf{73.0} & \textbf{73.1} & 50.1 & 52.3 & \textbf{66.8}/\textbf{57.6}/\textbf{65.5} \\
                   \midrule
                   & \checkmark & 2        & \underline{66.1} & \underline{72.4} & \textbf{44.7} &\textbf{ 47.3} & \underline{72.4} & \underline{72.7} & \underline{49.7} & \underline{53.1} & \underline{67.4}/\textbf{58.9}/\textbf{66.2} \\
                   & \checkmark & 4        & \textbf{66.3} & \textbf{72.5} & \underline{44.6} & \underline{47.1} & \textbf{72.5} & \textbf{72.9} & \textbf{50.0} & \textbf{53.7} & \textbf{67.6}/\underline{58.8}/\textbf{66.2} \\
                   & \checkmark & 8        & 66.0 & 72.3 & 44.3 & 46.6 & 72.1 & 72.7 & 49.0 & 52.0 & 67.3/57.5/65.7 \\
        \bottomrule
    \end{tabular}
    }
    \label{tab:ablation-sampler}
\end{table} 

\noindent\headbf{Hyper-parameters.} \tablename~\ref{tab:parameters} delineates the comprehensive hyperparameter configurations adopted for our two-stage training paradigm. During the agnostic segmentor training phase, the mask encoder is kept frozen; we optimize solely the mask decoder for a single epoch using a learning rate of $1 \times 10^{-4}$ and a gradient clipping maximum norm of $0.01$. In the unified joint training phase, optimization is extended across the projectors, LLM, mask encoder, mask decoder, and mask memory for one epoch. We employ LoRA~\cite{hu2022lora} for LLM fine-tuning, configured with a rank $r=128$ and a scaling factor $\alpha=256$. The learning rate is strictly set to $1 \times 10^{-5}$ for the mask encoder and $1 \times 10^{-4}$ for all other trainable modules, with the gradient clipping threshold relaxed to a maximum norm of $1.0$. Across both training stages, optimization is driven by AdamW~\cite{loshchilov2017adamw} with momentum parameters $(\beta_1, \beta_2) = (0.9, 0.999)$. The learning rate is governed by a cosine annealing schedule incorporating an initial warmup ratio of $0.03$. We apply a weight decay of $0.05$, configure the mask memory capacity to $8$, and enforce dataset-balanced resampling with a temperature $t=0.1$. Large-scale distributed training is executed on $32$ NVIDIA GPUs. For the agnostic stage, this hardware configuration yields an effective global batch size of $128$. For the unified joint training, we implement a modality-aware batching strategy with a per-device batch size of $1$; video data strictly maintains a global batch size of $32$, whereas image data applies an image batch multiplier of $4$, thereby achieving an effective global batch size of $128$. 

\begin{table*}[!t]
    \centering
    \setlength{\tabcolsep}{2pt}
    \caption{Comparison across image and video generic segmentation benchmarks.}
    \resizebox{\textwidth}{!}{
        \begin{tabular}{lcccccc}
            \toprule
            \multirow{3}{*}{\textbf{Method}}
             & \multicolumn{3}{c}{\textbf{I-Gen. Seg.}}
             & \multicolumn{3}{c}{\textbf{V-Gen. Seg.}}                                           \\
            \cmidrule(lr){2-4} \cmidrule(lr){5-7}
             & \textbf{\textit{COCO}}$_\textit{\textrm{Pan.}}$
             & \textbf{\textit{COCO}}$_\textit{\textrm{Sem.}}$
             & \textbf{\textit{COCO}}$_\textit{\textrm{Ins.}}$
             & \textbf{\textit{VIPSeg}}$_\textit{\textrm{Pan.}}$
             & \textbf{\textit{VSPW}}$_\textit{\textrm{Sem.}}$
             & \textbf{\textit{YT-VIS19}}$_\textit{\textrm{Ins.}}$                                \\
             & \scriptsize PQ/PQ$^\textrm{Th}$/PQ$^\textrm{St}$
             & \scriptsize mIoU
             & \scriptsize mAP
             & \scriptsize VPQ$^1$/VPQ$^2$/VPQ$^4$/VPQ$^6$/VPQ
             & \scriptsize mIoU/mVC$_8$/mVC$_{16}$
             & \scriptsize AP/AP50                                                                 \\
            \midrule
            \rowcolor{LightGray} \multicolumn{7}{l}{\textbf{\textit{Non-MLLM-Based Specialists}}} \\
            \color{gray} Mask2Former-L~\cite{cheng2022maskformer}
             & \color{gray} 57.8/64.2/48.1
             & \color{gray} 67.4
             & \color{gray} 48.6
             & \color{gray} \xmark
             & \color{gray} \xmark
             & \color{gray} \xmark                                                                \\
            \color{gray} VideoKNet~\cite{li2022videoknet}
             & \color{gray} \xmark
             & \color{gray} \xmark
             & \color{gray} \xmark
             & \color{gray} 43.3/40.5/38.3/37.2/39.8
             & \color{gray} 38.0/87.2/82.3
             & \color{gray} 54.1/79.0                                                             \\
            \midrule
            \rowcolor{LightGray} \multicolumn{7}{l}{\textbf{\textit{MLLM-Based Generalists}}}     \\
            PSALM~\cite{zhang2024psalm}
             & \textbf{55.9}/--/--
             & \textbf{66.6}
             & 45.7
             & \xmark
             & \xmark
             & \xmark                                                                             \\
            OMG-LLaVA~\cite{zhang2024omgllava}
             & 53.8/--/--
             & --
             & --
             & \underline{54.4}/\textbf{49.1}/\textbf{46.7}/\textbf{45.3}/\textbf{48.9}
             & --
             & --                                                                                 \\
            X-SAM~\cite{wang2026xsam}
             & \underline{54.7}/\textbf{60.7}/\textbf{45.7}
             & \underline{66.5}
             & \textbf{47.0}
             & \xmark
             & \xmark
             & \xmark                                                                             \\
            \rowcolor{LightGreen} \textbf{X2SAM}
             & 54.1/\underline{60.3}/\underline{44.9}
             & 64.8
             & \underline{45.8}
             & \textbf{59.3}/\underline{48.4}/\underline{42.4}/\underline{38.9}/\underline{47.3}
             & \textbf{65.1}/\textbf{90.0}/\textbf{86.5}
             & \textbf{69.9}/\textbf{88.4}                                                        \\
            \bottomrule
        \end{tabular}
    }
    \label{tab:gen-seg}
\end{table*} \section{More Ablation Studies}
\label{sec:app_ablation}
\noindent\headbf{Agnostic Segmentor Training.} 
\tablename~\ref{tab:seg-comparison} ablates the impact of training data sources on the agnostic segmentor. In image segmentation, although COCO performs competitively on RefCOCO due to domain alignment, scaling to SAM-Sub and ultimately SAM-1B consistently improves generalization. Training on SAM-1B achieves the best overall image-level results, notably peaking at 52.5 PQ in I-Gen. and 66.8 gIoU in I-Rea. Similarly, in video segmentation, while the intermediate SAM-Sub dataset exhibits slight drops in V-Gen. and V-OV compared to COCO, scaling to the massive SAM-1B dataset fully resolves this limitation. SAM-1B delivers the highest scores across all video tracks, yielding 45.4 PQ in V-Gen. and 59.5 mAP in V-OV. Overall, leveraging large-scale, diverse data like SAM-1B is essential for learning robust representations across universal spatio-temporal segmentation tasks.

\noindent\headbf{Region Sampler.} \tablename~\ref{tab:ablation-sampler} ablates the region sampler's feature source and spatial kernel size. First, utilizing features from the Mask Encoder consistently outperforms the Vision Encoder across most tasks, notably improving point-prompted I-Int from 64.7\% to 66.2\% mIoU. Second, applying a localized spatial kernel ($K=4$) yields optimal results compared to global aggregation ($\infty$) or larger kernels ($K=8$), achieving peak performance on I-Int and highly competitive scores on video benchmarks. Consequently, we adopt the Mask Encoder with a kernel size of $4$ as our default configuration.

\noindent\headbf{MLLMs.}
As shown in \tablename~\ref{tab:mllm}, Qwen3VL-4B achieves the best overall performance on image segmentation benchmarks, leading across I-OV, I-Ref., I-Rea., and I-VGD. Its strong results on referring and reasoning segmentation indicate better alignment between language instructions and spatial visual understanding.
In video segmentation, Qwen3VL-4B also performs best on most instruction-guided tasks, including V-Ref., V-Rea., and V-VGD, while matching the top YT19 result. The Siglip2-based models remain competitive in V-OV, with Siglip2+Phi3-3.8B achieving the highest score. Overall, Qwen3VL-4B provides the most balanced performance across image and video segmentation tasks.

\section{More Benchmark Results}
\label{sec:app_benchmark}
\begin{table*}[!t]
    \centering
    \setlength{\tabcolsep}{2pt}
    \caption{Comparison across image and video referring segmentation benchmarks.}
    \begin{tabular}{lccccc}
        \toprule
        \multirow{3}{*}{\textbf{Method}}
                                                            & \multicolumn{3}{c}{\textbf{I-Ref. Seg.}}
                                                            & \multicolumn{2}{c}{\textbf{V-Ref. Seg.}}                                                                                                                                                                                                                            \\
        \cmidrule(lr){2-4} \cmidrule(lr){5-6}
                                                            & \textbf{\textit{RefCOCO}}                          & \textbf{\textit{RefCOCO+}}                          & \textbf{\textit{RefCOCOg}}                          & \textbf{\textit{Ref-YT21-Val}}                    & \textbf{\textit{Ref-DV17-Val}}                    \\
                                                            & \scriptsize cIoU(val/testA/testB)                  & \scriptsize cIoU(val/testA/testB)                   & \scriptsize cIoU(val/test)                          & \scriptsize \mj/\mf/\mjf                              & \scriptsize \mj/\mf/\mjf                              \\
        \midrule
        \rowcolor{LightGray} \multicolumn{6}{l}{\textbf{\textit{Non-MLLM-Based Specialists}}}                                                                                                                                                                                                                                       \\
        \color{gray} CRIS-RN101~\cite{wang2022cris}         & \color{gray} 70.5/73.2/66.1                        & \color{gray} 62.3/68.1/53.7                         & \color{gray} 59.9/60.4                              & \color{gray} \xmark                               & \color{gray} \xmark                               \\
        \color{gray} ReferFormer-L~\cite{wu2022referformer} & \color{gray} \xmark                                & \color{gray} \xmark                                 & \color{gray} \xmark                                 & \color{gray} 62.3/66.2/64.2                       & \color{gray} 57.6/63.4/60.5                       \\
        \color{gray} UniRef++-L~\cite{wu2023uniref++}        & \color{gray} 79.1/82.2/77.5                        & \color{gray} 68.4/74.0/61.5                         & \color{gray} 71.4/72.8                              & \color{gray} 64.8/69.0/66.9                       & \color{gray} 63.4/70.9/67.2                       \\
        \midrule
        \rowcolor{LightGray} \multicolumn{6}{l}{\textbf{\textit{MLLM-Based Generalists}}}                                                                                                                                                                                                                                           \\
        LISA-7B~\cite{lai2024lisa}                          & 74.9/79.1/72.3                                     & 65.1/70.8/58.1                                      & 67.9/70.6                                           & 53.4/54.3/53.9                                    & 62.2/67.3/64.8                                    \\
        VISA-7B~\cite{yan2024visa}                          & 72.4/75.5/68.1                                     & 59.8/64.8/53.1                                      & 65.5/66.4                                           & 59.8/63.2/61.5                                    & 66.3/72.5/69.4            \\
        UniPixel-7B~\cite{liu2025unipixel}                  & 80.8/83.0/77.4                                     & 75.3/80.1/70.0                                      & 76.4/77.1                                           & \underline{69.5}/\underline{72.4}/\underline{71.0}& \underline{72.7}/\underline{80.1}/\underline{76.4}         \\
        HyperSeg~\cite{wei2024hyperseg}                     & \textbf{84.8}/\textbf{85.7}/\textbf{83.4}          & \textbf{79.0}/\textbf{83.5}/\textbf{75.2}           & \underline{79.4}/\underline{78.9}                   & {--}/{--}/68.5                                    & {--}/{--}/71.2                        \\
        \rowcolor{LightGreen} \textbf{X2SAM}                & \underline{84.0}/\underline{85.5}/\underline{81.8} & \underline{78.4}/\underline{82.4}/\underline{74.3}  & \textbf{81.9}/\textbf{83.2}                         & \textbf{76.0}/\textbf{80.9}/\textbf{78.5}         & \textbf{75.2}/\textbf{82.8}/\textbf{79.0}                                    \\
        \bottomrule
    \end{tabular}
    \label{tab:ref-seg}
\end{table*}

\begin{table*}[!t]
    \centering
    \setlength{\tabcolsep}{2pt}
    \caption{Comparison across image and video grounded conversation generation segmentation benchmarks.  \textcolor{gray}{Grayed} values means the method is reported in the original paper, * means the method is re-evaluated in this work.}
\begin{tabular}{lcccccccccccc}
            \toprule
            \multirow{3}{*}{\textbf{Method}}
                                                                    & \multicolumn{8}{c}{\textbf{I-GCG Seg.}}
                                                                    & \multicolumn{4}{c}{\textbf{V-GCG Seg.}}                                                                                                                                                          \\
            \cmidrule(lr){2-9} \cmidrule(lr){10-13}
                                                                    & \multicolumn{4}{c}{\textbf{\textit{GLaMM}}$_\textit{\textrm{Val}}$}
                                                                    & \multicolumn{4}{c}{\textbf{\textit{GLaMM}}$_\textit{\textrm{Test}}$}
                                                                    & \multicolumn{4}{c}{\textbf{\textit{V-GLaMM}}}                                                                                                                                                    \\
            \cmidrule(lr){2-5} \cmidrule(lr){6-9} \cmidrule(lr){10-13}
                                                                    & \scriptsize METEOR                                                   & \scriptsize CIDEr & \scriptsize AP50   & \scriptsize mIoU
                                                                    & \scriptsize METEOR                                                   & \scriptsize CIDEr & \scriptsize AP50   & \scriptsize mIoU
                                                                    & \scriptsize METEOR                                                   & \scriptsize CIDEr & \scriptsize Recall & \scriptsize mIoU                                                                 \\
            \midrule
            \rowcolor{LightGray} \multicolumn{13}{l}{\textbf{\textit{MLLM-Based Image Generalists}}}                                                                                                                                                                   \\
            LISA-7B~\cite{lai2024lisa}                              & 13.0                                                                 & 33.9              & 25.2               & 62.0             & 12.9   & 32.2   & 24.8   & 61.7   & \xmark & \xmark & \xmark & \xmark \\
            GLaMM-7B~\cite{rasheed2024glamm}                        & \underline{15.2}                                                                 & \underline{43.1}              & 28.9               & 65.8             & 14.6   & \underline{37.9}   & 27.2   & 64.6   & \xmark & \xmark & \xmark & \xmark \\
            X-SAM~\cite{wang2026xsam}                               & \textbf{15.4}                                                                 & \textbf{46.3}              & \textbf{33.2}               & \textbf{69.4}             & \textbf{15.1}   & \textbf{42.7}   & \textbf{32.9}   & \textbf{69.0}   & \xmark & \xmark & \xmark & \xmark \\
            \midrule
            \rowcolor{LightGray} \multicolumn{13}{l}{\textbf{\textit{MLLM-Based Video Generalists}}}                                                                                                                                                                   \\
            PG-Video-LLaVA~\cite{munasinghe2023pgvllava}            & \xmark                                                               & \xmark            & \xmark             & \xmark           & \xmark & \xmark & \xmark & \xmark & \color{gray} 10.0 & \color{gray} 1.0  & \color{gray} 9.3  & \color{gray} 24.0 \\
            GLaMM~\cite{rasheed2024glamm}~+SAM2~\cite{ravi2024sam2} & \xmark                                                               & \xmark            & \xmark             & \xmark           & \xmark & \xmark & \xmark & \xmark & \color{gray} 9.7  & \color{gray} 15.0 & \color{gray} 11.7 & \color{gray} 28.6 \\
            Video-GLaMM~\cite{munasinghe2024videoglamm}             & \xmark                                                               & \xmark            & \xmark             & \xmark           & \xmark & \xmark & \xmark & \xmark & \color{gray} 10.3 & \color{gray} 59.0 & \color{gray} 37.5 & \color{gray} 62.3 \\
            Video-GLaMM*~\cite{munasinghe2024videoglamm}            & \xmark                                                               & \xmark            & \xmark             & \xmark           & \xmark & \xmark & \xmark & \xmark & \underline{7.4}  & \underline{19.5} & \underline{30.2} & \underline{54.3} \\
            \rowcolor{LightGreen} \textbf{X2SAM}                     & \underline{15.2}                                                                 & 35.6              & \underline{33.1}               & \underline{67.1}             & \underline{14.8}   & 33.6   & \underline{31.3}   & \underline{65.2}   &  \textbf{16.6}    &  \textbf{43.2}    &   \textbf{42.0}   & \textbf{75.8} \\
            \bottomrule
        \end{tabular}
\label{tab:gcg-seg}
\end{table*} 

\begin{table*}[!t]
    \centering
    \setlength{\tabcolsep}{2pt}
    \caption{Comparison on object-centric segmentation tasks, including image interactive segmentation (I-Int.) and video object segmentation (V-Obj.) benchmarks.}
    \resizebox{\textwidth}{!}{
        \begin{tabular}{lccccccc}
            \toprule
            \multirow{3}{*}{\textbf{Method}}
                                                      & \multicolumn{4}{c}{\textbf{I-Int. Seg.}}
                                                      & \multicolumn{3}{c}{\textbf{V-Obj. Seg.}}                                                                                                                                                                                                                                                                                \\
            \cmidrule(lr){2-5} \cmidrule(lr){6-8}
                                                      & \textbf{\textit{COCO}}$_\textit{\textrm{Point}}$ & \textbf{\textit{COCO}}$_\textit{\textrm{Scribble}}$ & \textbf{\textit{COCO}}$_\textit{\textrm{Box}}$ & \textbf{\textit{COCO}}$_\textit{\textrm{Mask}}$ & \textbf{\textit{YT-VOS19}}$_\textit{\textrm{Seen}}$ & \textbf{\textit{YT-VOS19}}$_\textit{\textrm{UnSeen}}$ & \textbf{\textit{YT-VOS19}}$_\textit{\textrm{All}}$\\
                                                      & \scriptsize mIoU/cIoU                            & \scriptsize mIoU/cIoU                               & \scriptsize mIoU/cIoU                          & \scriptsize mIoU/cIoU                           & \scriptsize \mj/\mf/\mjf                            & \scriptsize \mj/\mf/\mjf & \scriptsize \mj/\mf/\mjf                            \\
            \midrule
            \rowcolor{LightGray} \multicolumn{8}{l}{\textbf{\textit{Non-MLLM-Based Specialists}}}                                                                                                                                                                                                                                                                               \\
            \color{gray} SAM-L~\cite{kirillov2023sam} & \color{gray} 51.8/37.7                           & \color{gray} --                                     & \color{gray} 76.6/71.6                         & \color{gray} --                                 & \color{gray} --                                     & \color{gray} --        & \color{gray} --                              \\
            \color{gray} SAM2-H~\cite{ravi2024sam2}   & \color{gray} --                                  & \color{gray} --                                     & \color{gray} --                                & \color{gray} --                                 & \color{gray} 86.5/91.0/88.8                         & \color{gray} 84.7/92.8/88.8     & \color{gray} --                        \\
\midrule
            \rowcolor{LightGray} \multicolumn{8}{l}{\textbf{\textit{MLLM-Based Generalists}}}                                                                                                                                                                                                                                                                                   \\
            PSALM~\cite{zhang2024psalm}               & 64.3/\textbf{74.0}                                        & \textbf{66.9}/\textbf{80.0}                                           & 67.3/\textbf{80.9}                                      & 67.6/\textbf{82.4}                                       & \xmark                                                  & \xmark            & \xmark                                        \\
            X-SAM~\cite{wang2026xsam}                 & \underline{65.4}/62.9                                        & \textbf{66.9}/75.7                                           & \underline{69.6}/75.4                                      & \underline{69.7}/77.0                                       & \xmark                                                  & \xmark            & \xmark                                        \\
            \rowcolor{LightGreen} \textbf{X2SAM}       &  \textbf{67.7}/\underline{67.1}                                                &    \underline{66.3}/\underline{69.2}                                                 &   \textbf{70.3}/\underline{75.7}                                             &    \textbf{71.6}/\underline{81.5}                                             &     \textbf{74.3}/\textbf{77.7}/\textbf{76.0}                                                &     \textbf{58.6}/\textbf{64.1}/\textbf{61.4}    & \textbf{72.2}/\textbf{75.9}/\textbf{74.0}                                              \\
            \bottomrule
        \end{tabular}
    }
    \label{tab:obj-seg}
\end{table*}

\begin{table*}[ht]
    \centering
    \setlength{\tabcolsep}{2pt}
    \caption{Comparison across image chat benchmarks.}
\begin{tabular}{lccccc}
        \toprule
        \multirow{3}{*}{\textbf{Method}}
         & \multicolumn{5}{c}{\textbf{I-Chat}}                                      \\
        \cmidrule(lr){2-6}
         & \textbf{\textit{MME}}
         & \textbf{\textit{MMBench}}
         & \textbf{\textit{SEED-Bench}}
         & \textbf{\textit{POPE}}
         & \textbf{\textit{AI2D}}                                                   \\
         & \scriptsize Acc.
         & \scriptsize Acc.
         & \scriptsize Acc.
         & \scriptsize Acc.
         & \scriptsize Acc.                                                         \\
        \midrule
        \rowcolor{LightGray} \multicolumn{6}{l}{\textbf{\textit{Chat-based MLLMs}}} \\
        \color{gray} LLaVA-1.5~\cite{liu2024llava1x5}
         & \color{gray} 1510/--
         & \color{gray} 64.3
         & \color{gray} 58.6
         & \color{gray} 87.3
         & --                                                                       \\
        \color{gray} LLaVA-OV~\cite{li2024llavaov}
         & \color{gray} 1580/418
         & \color{gray} 80.8
         & \color{gray} 75.4
         & \color{gray} --
         & \color{gray} 81.4                                                        \\
        \color{gray} Qwen3-VL~\cite{qwen3vl}
         & \color{gray} --
         & \color{gray} 83.9
         & \color{gray} --
         & \color{gray} --
         & \color{gray} 84.1                                                        \\
        \midrule
        \rowcolor{LightGray} \multicolumn{6}{l}{\textbf{\textit{Seg.-based MLLMs}}} \\
        LISA~\cite{lai2024lisa}
         & 1/--
         & 0.4
         & --
         & 0
         & 0                                                                        \\
        PixelLM~\cite{zhang2024psalm}
         & 309/135
         & 17.4
         & --
         & 0
         & 0                                                                        \\
        GLaMM~\cite{rasheed2024glamm}
         & 14/--
         & 36.8
         & --
         & 0.94
         & 28.2                                                                     \\
        OMG-LLaVA~\cite{zhang2024omgllava}
         & 1177/235
         & 47.9
         & 56.5
         & 80.0
         & 42.9                                                                     \\
        X-SAM~\cite{wang2026xsam}
         & 1374/312
         & 69.3
         & 69.3
         & \textbf{89.3}
         & 62.6                                                                     \\
        \rowcolor{LightGreen} \textbf{X2SAM}
         & \textbf{1701}/\textbf{601}
         & \textbf{83.5}
         & \textbf{76.0}
         & \underline{88.2}
         & \textbf{82.0}                                                                          \\
        \bottomrule
    \end{tabular}
\label{tab:image-chat}
\end{table*}

\begin{table*}[ht]
    \centering
    \setlength{\tabcolsep}{2pt}
    \caption{Comparison across video chat benchmarks.}
\begin{tabular}{lcccc}
        \toprule
        \multirow{3}{*}{\textbf{Method}}
         & \multicolumn{4}{c}{\textbf{V-Chat}}                                      \\
        \cmidrule(lr){2-5}
         & \textbf{\textit{VideoMME}}
         & \textbf{\textit{MVBench}}
         & \textbf{\textit{MLVU}}
         & \textbf{\textit{LongVideoBench}}                                         \\
         & \scriptsize Acc.
         & \scriptsize Acc.
         & \scriptsize Acc.
         & \scriptsize Acc.                                                         \\
        \midrule
        \rowcolor{LightGray} \multicolumn{5}{l}{\textbf{\textit{Chat-based MLLMs}}} \\
        \color{gray} Video-LLaVA~\cite{munasinghe2023pgvllava}
         & \color{gray} 41.6
         & \color{gray} --
         & \color{gray} 29.3
         & \color{gray} 39.1                                                        \\
        \color{gray} VideoChat2~\cite{li2025videochat}
         & \color{gray} 43.8
         & \color{gray} 62.3
         & \color{gray} 37.4
         & \color{gray} 39.3                                                        \\
        \color{gray} Chat-UniVi-V~\cite{jin2024chatunivi}
         & \color{gray} 45.9
         & \color{gray} --
         & \color{gray} --
         & \color{gray} --                                                          \\
        \color{gray} VILA-1.5~\cite{lin2024vila}
         & \color{gray} 59.4
         & \color{gray} --
         & \color{gray} --
         & \color{gray} --                                                          \\
        \color{gray} Qwen3-VL~\cite{qwen3vl}
         & \color{gray} --
         & \color{gray} 68.9
         & \color{gray} 75.3
         & \color{gray} --                                                          \\
        \midrule
        \rowcolor{LightGray} \multicolumn{5}{l}{\textbf{\textit{Seg.-based MLLMs}}} \\
        \rowcolor{LightGreen} \textbf{X2SAM}
         & \textbf{74.4}
         & \textbf{63.1}
         & \textbf{67.1}
         & \textbf{57.4}                                                                         \\
        \bottomrule
    \end{tabular}
\label{tab:video-chat}
\end{table*}
 
\noindent\headbf{Generic Segmentation.} \tablename~\ref{tab:gen-seg} compares generic segmentation performance across image and video domains. In image generic segmentation (I-Gen. Seg.), X2SAM achieves 54.1/60.3/44.9 PQ/PQ$^\textrm{Th}$/PQ$^\textrm{St}$ on COCO panoptic segmentation, 45.8 mAP on COCO instance segmentation, and 64.8 mIoU on COCO semantic segmentation. These results show that X2SAM remains competitive with strong MLLM-based generalists such as PSALM and X-SAM.
In video generic segmentation (V-Gen. Seg.), X2SAM shows clear advantages. It achieves state-of-the-art results on VSPW, with 65.1 mIoU and 90.0/86.5 mVC$8$/mVC${16}$, and on YT-VIS19, with 69.9/88.4 AP/AP50. On VIPSeg, X2SAM obtains a competitive overall VPQ of 47.3 and the best VPQ$^1$ of 59.3 among all compared methods. Overall, these results show that X2SAM preserves strong image-level understanding while substantially improving performance on challenging video generic segmentation tasks.

\noindent\headbf{Referring Segmentation.} \tablename~\ref{tab:ref-seg} evaluates referring segmentation performance across both image and video domains. In the image domain (I-Ref. Seg.), X2SAM demonstrates highly competitive capabilities among MLLM-based generalists. Notably, it establishes a new state-of-the-art on the RefCOCOg benchmark, achieving top cIoU scores of 81.9 and 83.2 on the \textit{val} and \textit{test} splits, respectively. On the RefCOCO and RefCOCO+ datasets, X2SAM secures the second-best performance, closely trailing HyperSeg with competitive cIoU scores (e.g., 84.0 and 78.4 on their respective \textit{val} splits). Furthermore, X2SAM exhibits exceptional proficiency in the video domain (V-Ref. Seg.), where it significantly outperforms all evaluated methods. It attains the highest J\&F scores of 78.5 on Ref-YT21-Val and 79.0 on Ref-DV17-Val, surpassing the previously leading UniPixel-7B by substantial margins (+7.5 and +2.6 absolute improvements in J\&F, respectively). Overall, these results underscore X2SAM's robust and unified architecture, demonstrating superior temporal comprehension in complex video sequences alongside top-tier spatial grounding in static images.

\noindent\headbf{GCG Segmentation.} \tablename~\ref{tab:gcg-seg} evaluates Grounded Conversation Generation (GCG) segmentation in both image and video domains. For image GCG segmentation (I-GCG Seg.), X2SAM shows strong spatial reasoning on \textit{GLaMM}$\textit{\textrm{Val}}$ and \textit{GLaMM}$\textit{\textrm{Test}}$. It achieves 33.1 AP50 and 67.1 mIoU on the validation set, and 31.3 AP50 and 65.2 mIoU on the test set. These results surpass multimodal generalists such as LISA-7B and GLaMM-7B, while approaching the image-specialized X-SAM model. For video GCG segmentation (V-GCG Seg.), X2SAM achieves strong results on \textit{V-GLaMM} under our evaluation protocol, with leading METEOR, Recall, and mIoU among the compared video generalists (16.6 METEOR, 43.2 CIDEr, 42.0 Recall, and 75.8 mIoU). This substantially improves over video generalist baselines, including both the original and recently re-evaluated Video-GLaMM results. Overall, these results show that X2SAM can generate grounded descriptions with accurate object segmentation across both image and video modalities.

\noindent\headbf{Object-Centric Segmentation.} \tablename~\ref{tab:obj-seg} reports object-centric segmentation results, where I-Int. denotes image interactive segmentation and V-Obj. denotes video object segmentation. For image segmentation on COCO, X2SAM achieves the best performance among MLLM-based generalists across point, box, and mask prompts, with mIoU scores of 67.7, 70.3, and 71.6, respectively. Its scribble-prompt result (66.3 mIoU) and overall cIoU are also competitive, though comparable to or slightly lower than PSALM~\cite{zhang2024psalm}. For video segmentation on YT-VOS19, X2SAM shows strong generalization ability. Unlike MLLM baselines such as PSALM and X-SAM~\cite{wang2026xsam}, which are limited to image tasks, X2SAM extends to video object segmentation and achieves \mj/\mf/\mjf scores of 72.2/75.9/74.0. While it still lags behind specialized non-MLLM video models such as SAM2-H~\cite{ravi2024sam2}, these results demonstrate its versatility across multimodal object-centric segmentation tasks without task-specific architectural designs.

\noindent\headbf{Image Chat.} \tablename~\ref{tab:image-chat} evaluates image chat capabilities across five benchmarks. X2SAM achieves state-of-the-art performance among segmentation-based MLLMs on nearly all metrics. Specifically, X2SAM attains 1701/601 on MME~\cite{fu2024mme}, 83.5 on MMBench~\cite{liu2024mmbench}, 76.0 on SEED-Bench~\cite{li2024seed}, and 82.0 on AI2D~\cite{kembhavi2016ai2d}, outperforming X-SAM~\cite{wang2026xsam} and OMG-LLaVA~\cite{zhang2024omgllava}. While X-SAM retains the highest POPE~\cite{li2023pope} score (89.3), X2SAM remains competitive (88.2). Furthermore, X2SAM's chat performance rivals, and sometimes exceeds, chat-specialized models like LLaVA-OV~\cite{li2024llavaov} (e.g., 83.5 vs. 80.8 on MMBench). These findings suggest that fine-grained segmentation may undermine multimodal understanding and reasoning abilities.

\noindent\headbf{Video Chat.} \tablename~\ref{tab:video-chat} compares video chat performance across four benchmarks. As the only segmentation-based MLLM evaluated, X2SAM achieves strong accuracy: 74.4\%, 63.1\%, 67.1\%, and 57.4\% on VideoMME~\cite{fu2025videomme}, MVBench~\cite{li2024mvbench}, MLVU~\cite{zhou2025mlvu}, and LongVideoBench~\cite{wu2024longvideobench}, respectively. It outperforms many chat-centric video MLLMs, including Video-LLaVA~\cite{munasinghe2023pgvllava}, VideoChat2~\cite{li2025videochat}, Chat-UniVi-V~\cite{jin2024chatunivi}, and VILA-1.5~\cite{lin2024vila}. Compared with Qwen3-VL~\cite{qwen3vl}, results are mixed, reflecting the trade-off of extending a chat-focused backbone with dense segmentation. Overall, X2SAM retains competitive video-chat ability while enabling pixel-level segmentation.

\section{More Visualization Results}
\label{sec:app_visualization}
\begin{figure}[!ht]
    \centering
    \includegraphics[width=0.76\linewidth]{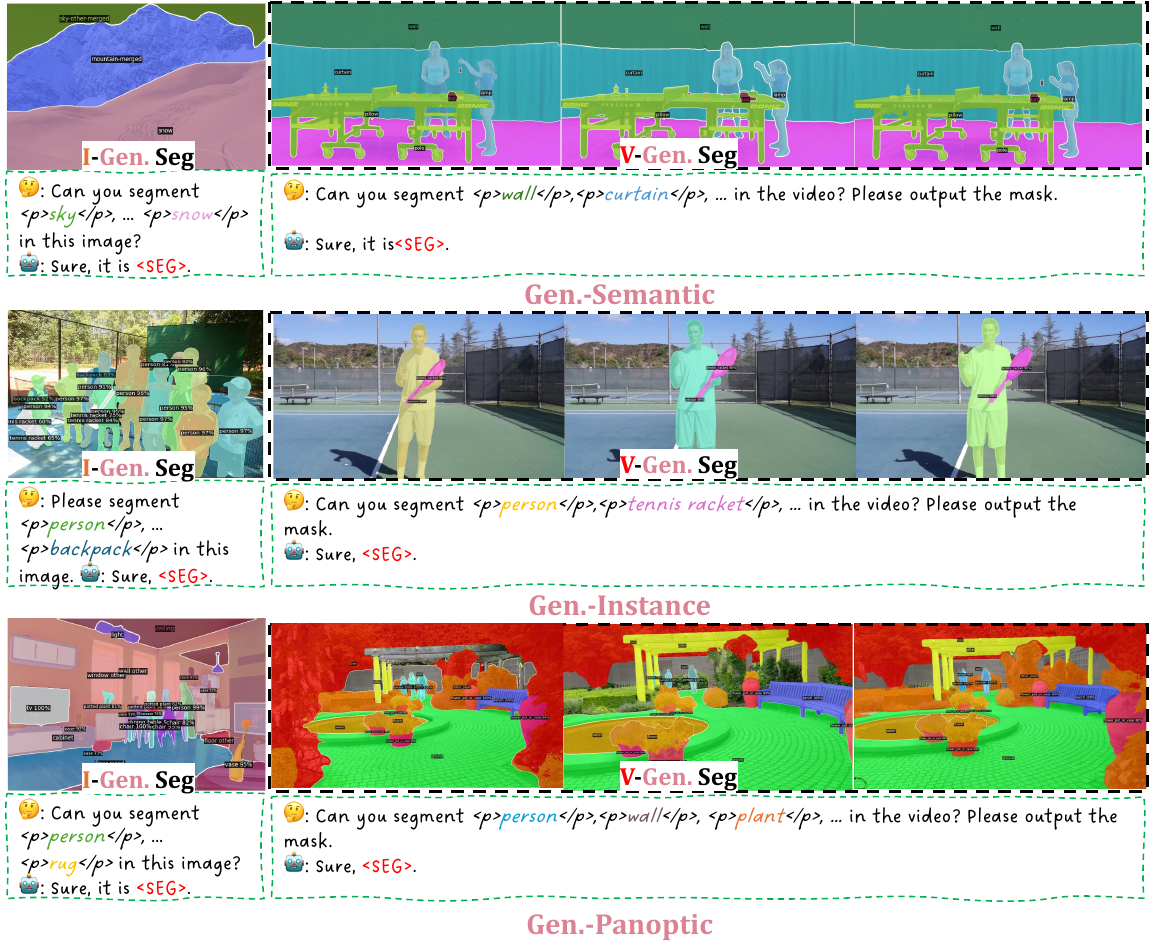}
    \caption{Visualization results of generic segmentation across images and videos, including semantic, instance, and panoptic segmentation.}
    \label{fig:genseg}
\end{figure}

\begin{figure}[!ht]
    \centering
    \includegraphics[width=0.76\linewidth]{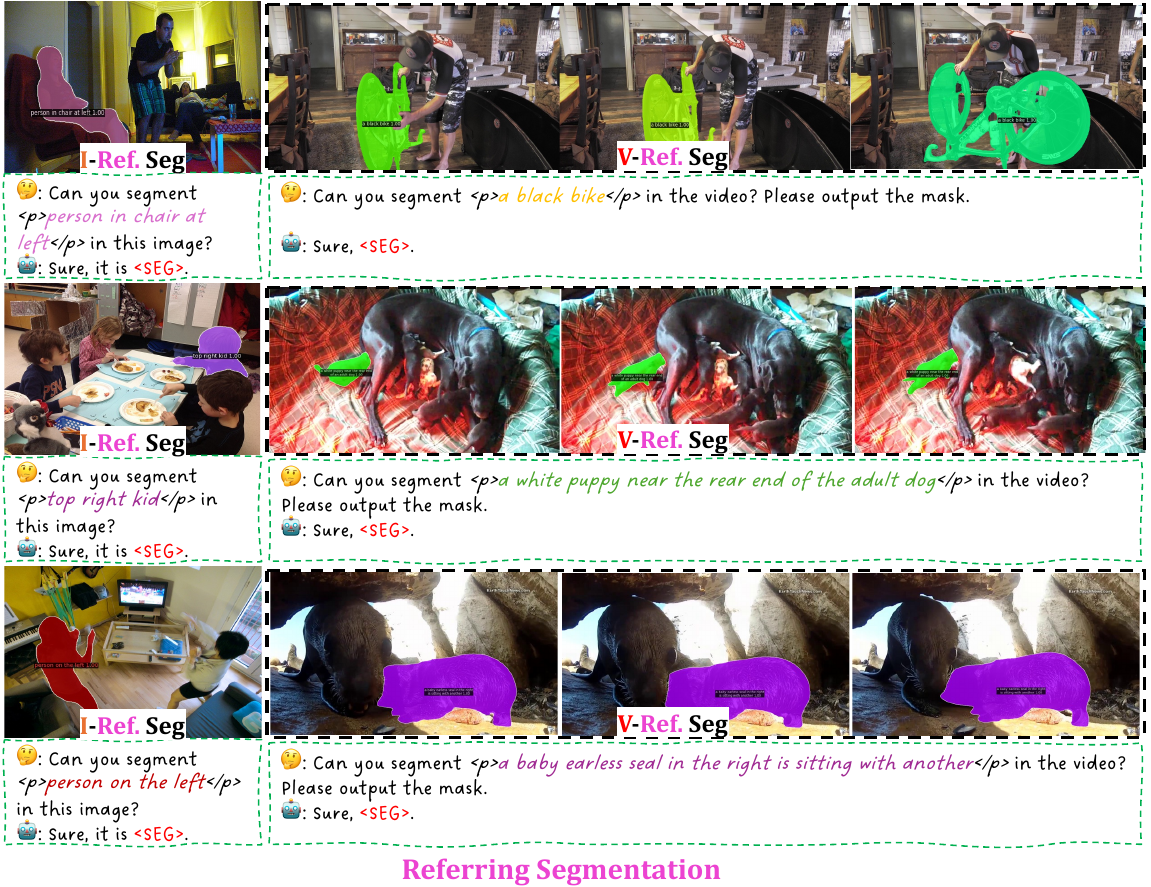}
    \caption{Visualization results of referring segmentation across images and videos.}
    \label{fig:refseg}
\end{figure}

\begin{figure}[!ht]
    \centering
    \includegraphics[width=0.76\linewidth]{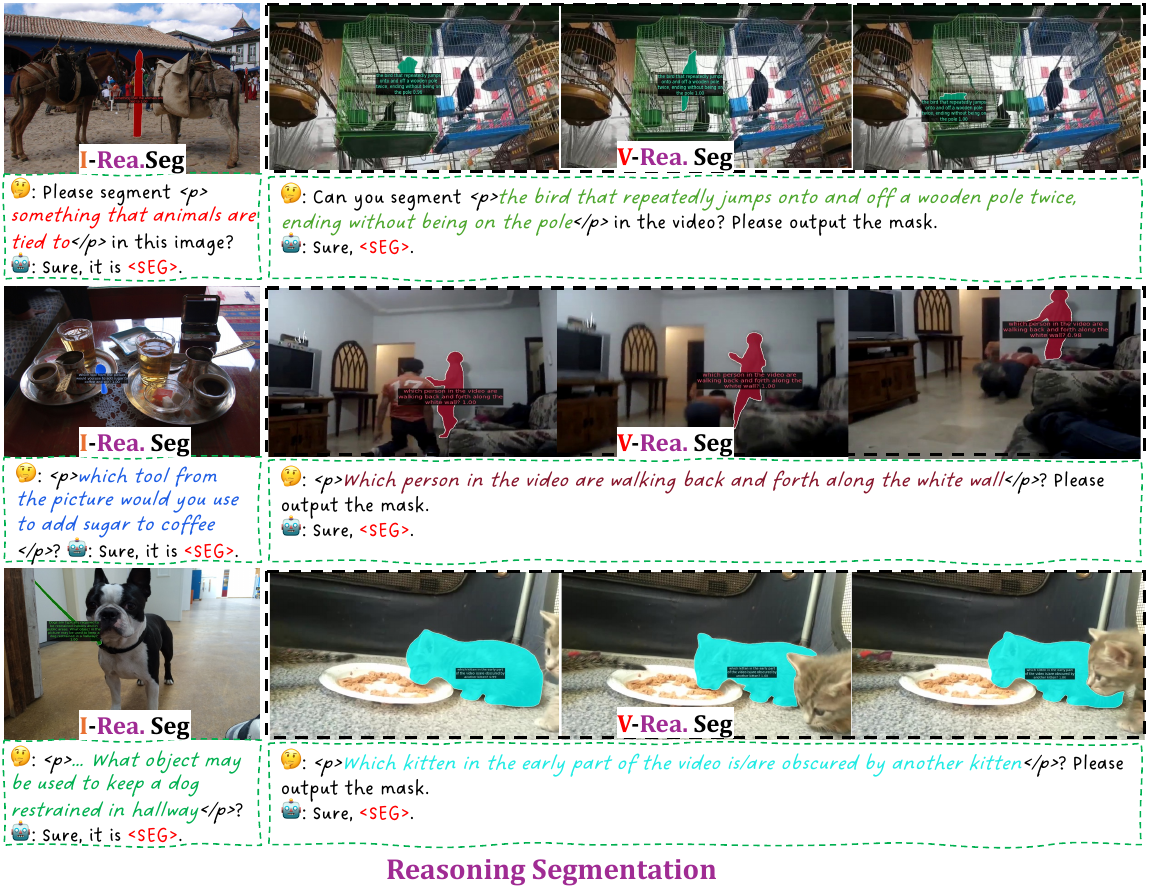}
    \caption{Visualization results of reasoning segmentation across images and videos.}
    \label{fig:reaseg}
\end{figure}

\begin{figure}[!ht]
    \centering
    \includegraphics[width=0.76\linewidth]{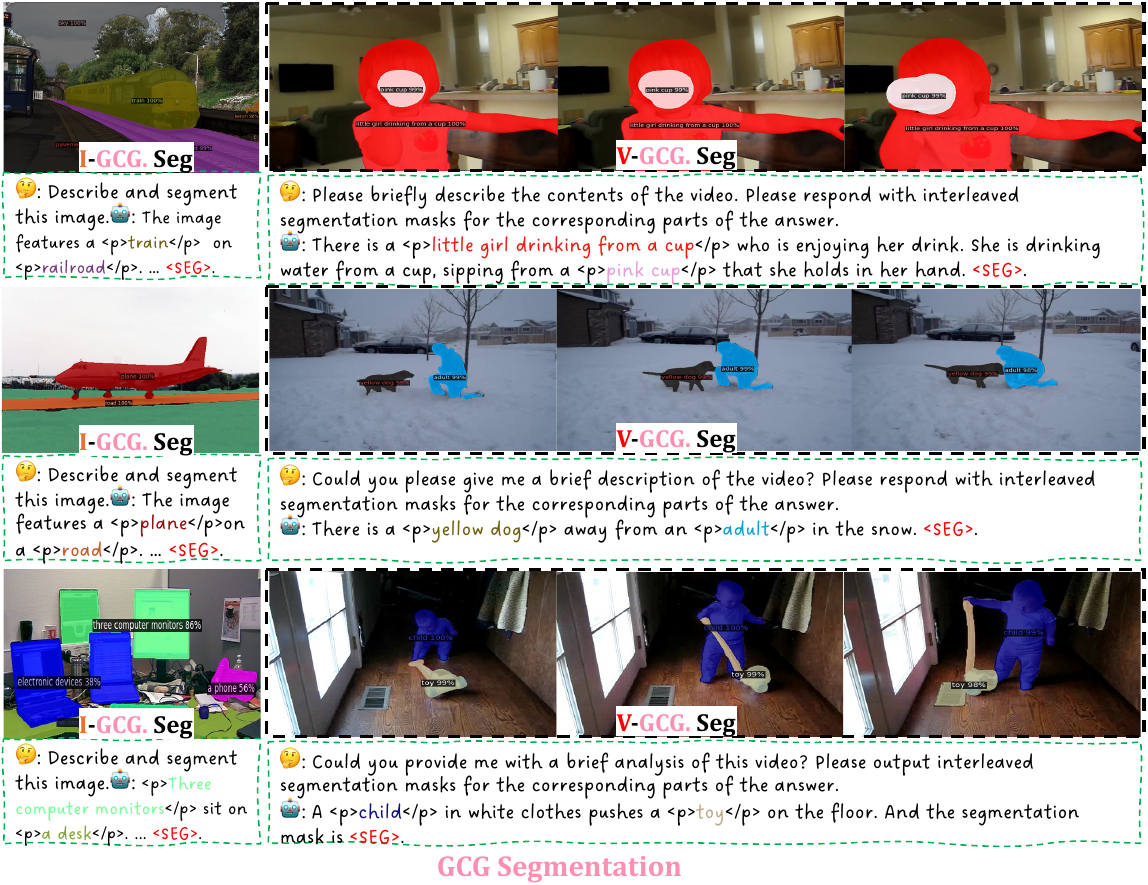}
    \caption{Visualization results of grounded conversation generation segmentation across images and videos.}
    \label{fig:gcgseg}
\end{figure}

\begin{figure}[!ht]
    \centering
    \includegraphics[width=0.76\linewidth]{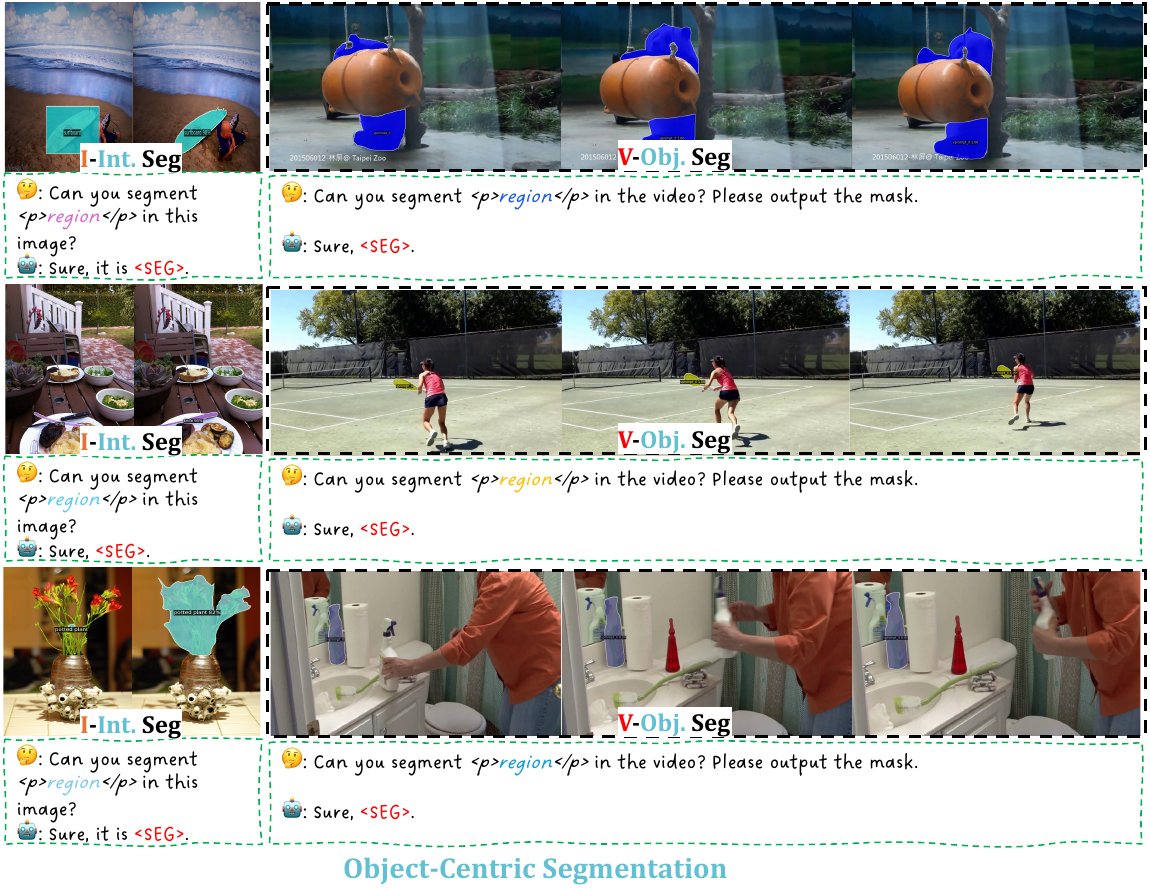}
    \caption{Visualization results of object-centric segmentation across images and videos, including image interactive segmentation (I-Int.) and video object segmentation (V-Obj.).}
    \label{fig:objseg}
\end{figure}

\begin{figure}[!ht]
    \centering
    \includegraphics[width=0.76\linewidth]{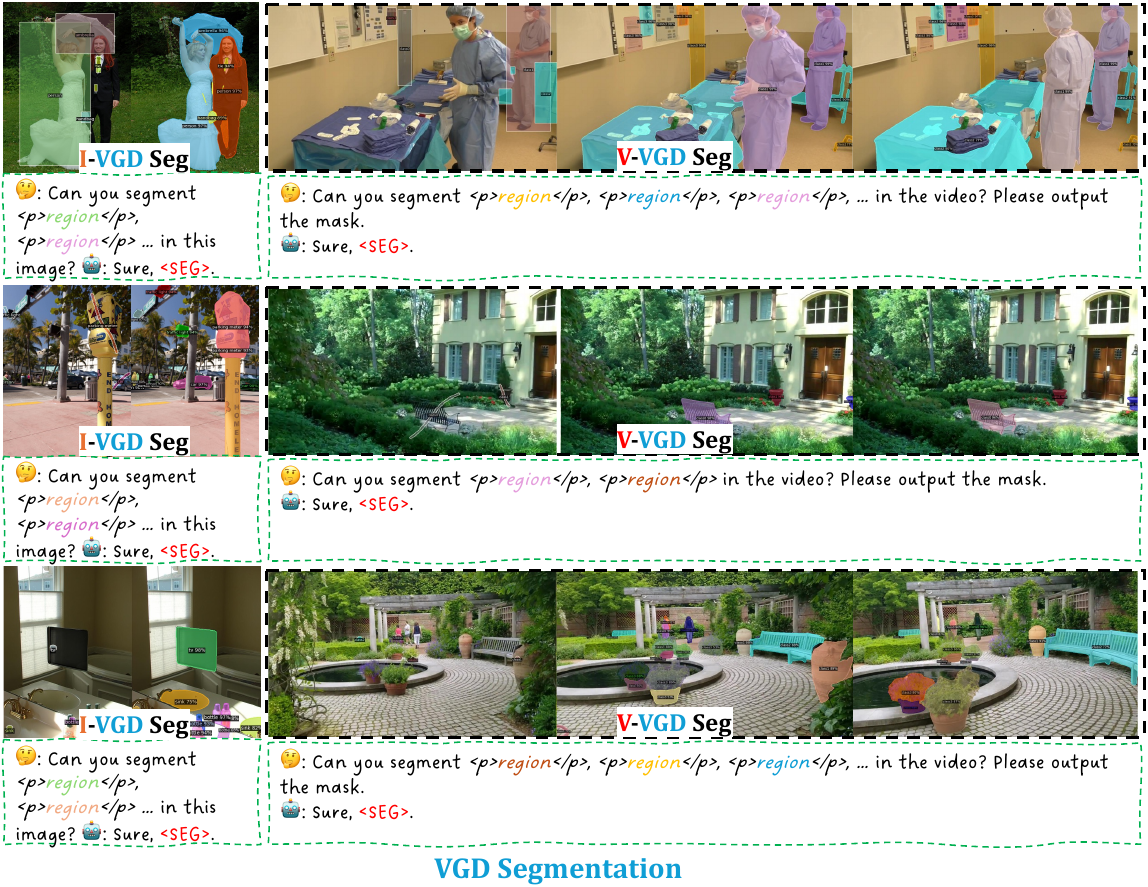}
    \caption{Visualization results of visual grounded segmentation across images and videos.}
    \label{fig:vgdseg}
\end{figure}

\begin{figure}[!ht]
    \centering
    \includegraphics[width=0.76\linewidth]{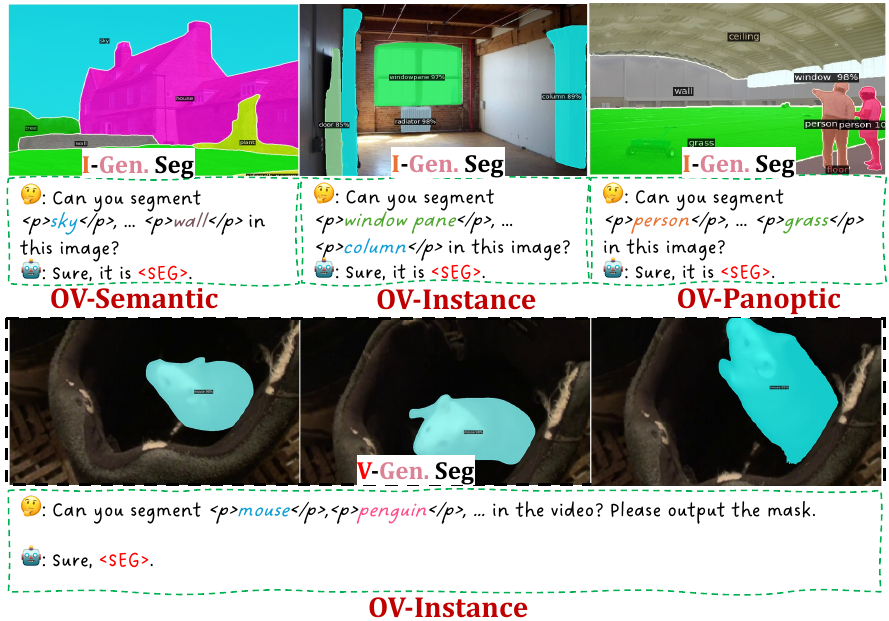}
    \caption{Visualization results of open-vocabulary segmentation across images and videos, including semantic, instance, and panoptic segmentation.}
    \label{fig:ovseg}
\end{figure}

\begin{figure}[!ht]
    \centering
    \includegraphics[width=0.88\linewidth]{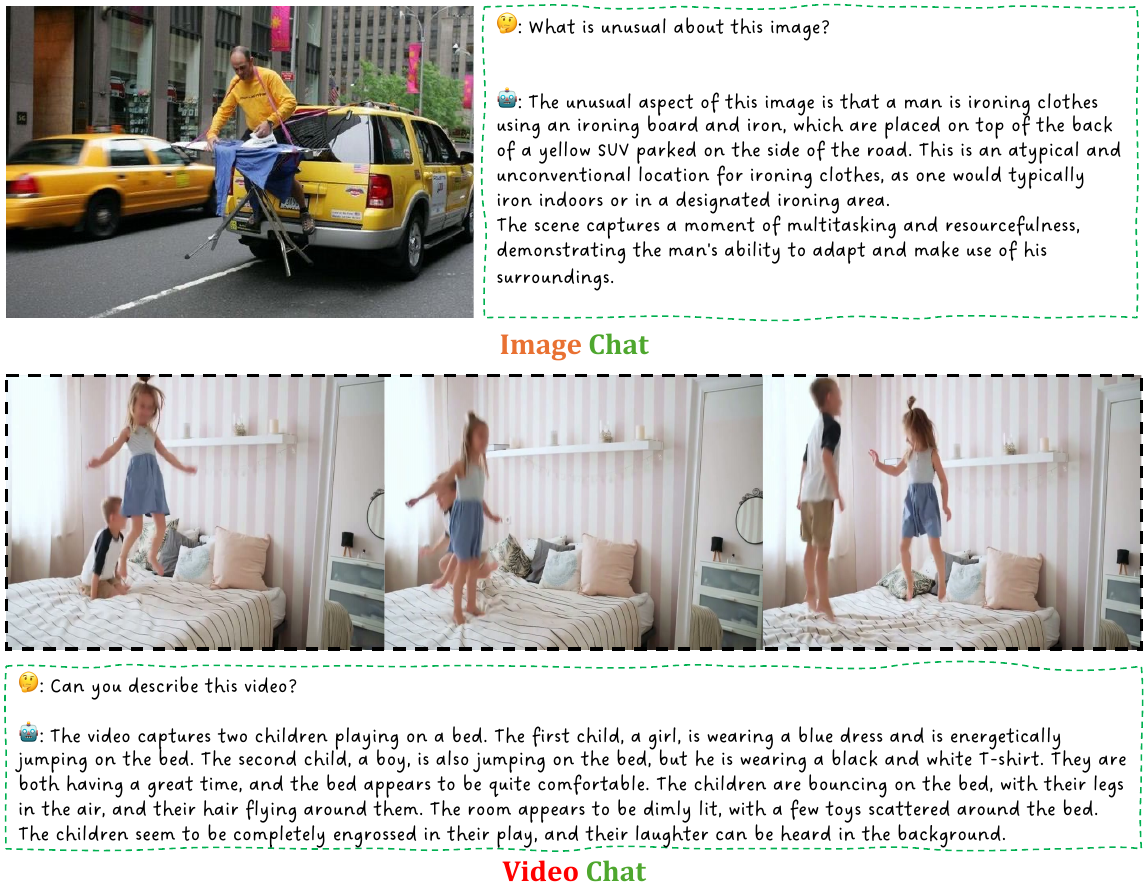}
    \caption{Visualization results of visual chat across images and videos.}
    \label{fig:chat}
\end{figure}

\figurename~\ref{fig:genseg}, \figurename~\ref{fig:refseg}, \figurename~\ref{fig:reaseg}, \figurename~\ref{fig:gcgseg}, \figurename~\ref{fig:objseg}, \figurename~\ref{fig:vgdseg}, \figurename~\ref{fig:ovseg} present additional visualization results of X2SAM on diverse image and video segmentation tasks, including generic, referring, reasoning, grounded conversation generation, object-centric, visual grounded, and open-vocabulary segmentation. These examples further demonstrate the model's ability to produce accurate and coherent masks under varied prompts, categories, and visual scenarios. \figurename~\ref{fig:chat} provides additional visual chat examples across images and videos.
\clearpage
         
\clearpage
\bibliographystyle{unsrtnat}
\bibliography{main}
\end{document}